\documentclass{article}

\usepackage{microtype}
\usepackage{graphicx}
\usepackage{booktabs} 

\usepackage{subcaption}
\usepackage{bm} 
\usepackage{natbib} 
\usepackage{multirow} 
\usepackage{makecell} 
\usepackage{color}
\usepackage{algorithm} 
\usepackage{marvosym} 

\usepackage{hyperref}

\usepackage[accepted]{icml2025}

\usepackage{amsmath}
\usepackage{amssymb}
\usepackage{mathtools}
\usepackage{amsthm}

\usepackage[capitalize,noabbrev]{cleveref}

\theoremstyle{plain}

\theoremstyle{definition}

\theoremstyle{remark}

\usepackage[textsize=tiny]{todonotes}

\icmltitlerunning{{ BOPO: Neural Combinatorial Optimization via Best-anchored and Objective-guided Preference Optimization}}

\begin{document}

\twocolumn[
\icmltitle{{ BOPO: Neural Combinatorial Optimization via \\ Best-anchored and Objective-guided Preference Optimization}}



\icmlsetsymbol{equal}{*}

\begin{icmlauthorlist}
\icmlauthor{Zijun Liao}{equal,sch}
\icmlauthor{Jinbiao Chen}{equal,sch}
\icmlauthor{Debing Wang}{sch}
\icmlauthor{Zizhen Zhang$^\dagger$}{sch}
\icmlauthor{Jiahai Wang}{sch}
\end{icmlauthorlist}

\icmlaffiliation{sch}{School of Computer Science and Engineering, Sun Yat-sen University, China}

\icmlcorrespondingauthor{Zizhen Zhang}{zhangzzh7@mail.sysu.edu.cn}

\icmlkeywords{Neural Combinatorial Optimization, Preference Optimization, Machine Learning}

\vskip 0.3in
]



\printAffiliationsAndNotice{\icmlEqualContribution} 

\begin{abstract}
Neural Combinatorial Optimization (NCO) has emerged as a promising approach for NP-hard problems. However, prevailing RL-based methods suffer from low sample efficiency due to sparse rewards and underused solutions. We propose { \emph{Best-anchored and Objective-guided Preference Optimization (BOPO)}}, a training paradigm that leverages solution preferences via objective values. It introduces: (1) a { best-anchored} preference pair construction for better explore and exploit solutions, and (2) an { objective-guided pairwise} loss function that adaptively scales gradients via objective differences, removing reliance on reward models or reference policies. Experiments on Job-shop Scheduling Problem (JSP), Traveling Salesman Problem (TSP), and Flexible Job-shop Scheduling Problem (FJSP) show { BOPO} outperforms state-of-the-art neural methods, reducing optimality gaps impressively with efficient inference. { BOPO} is architecture-agnostic, enabling seamless integration with existing NCO models, and establishes preference optimization as a principled framework for combinatorial optimization.

\end{abstract}


\section{Introduction}
\label{sec:introduction}

Combinatorial optimization problems (COPs), such as scheduling \cite{jsp_review,jsp_review2} and routing problems \cite{vidal2020concise,berghman2023review}, are widely applied in real-world scenarios and have attracted significant research attention. Most COPs are NP-hard, making them challenging to find optimal solutions. Exact methods, such as branch-and-bound algorithms, require exponential computation time as the problem size increases. Consequently, heuristic methods have proven effective in obtaining high-quality solutions within reasonable time over the past decades. Nevertheless, these methods still heavily rely on expert knowledge and extensive iterative search.

In the emerging field of neural combinatorial optimization (NCO), deep neural models are employed to automatically learn heuristics from training data, enabling the rapid construction of high-quality solutions in an end-to-end fashion \cite{maz21,ben21,yan22,jsp_rl_review,zha23b,gar24}. Early research \cite{pn} adopted supervised learning (SL) to train the deep models, which required (near-) optimal solutions produced by expensive specialized solvers as labels. Different from SL, reinforcement learning (RL), which does not require labels, has emerged as the mainstream training paradigm for NCO \cite{tsp_am,tsp_pomo,l2d,rs}. However, RL encounters challenges such as sparse rewards and low sample efficiency \cite{drl1}. Recently, self-labeling learning (SLL) \cite{slim,sil} was proposed to partially address these issues by sampling multiple solutions and treating the best one among them as a pseudo-label for model training. Nevertheless, SLL still faces the challenge of low sample efficiency, as all sampled solutions except the optimal one are discarded during training.

\begin{figure*}[t]
\vskip 0.2in
\centering
\centerline{\includegraphics[scale=0.7]{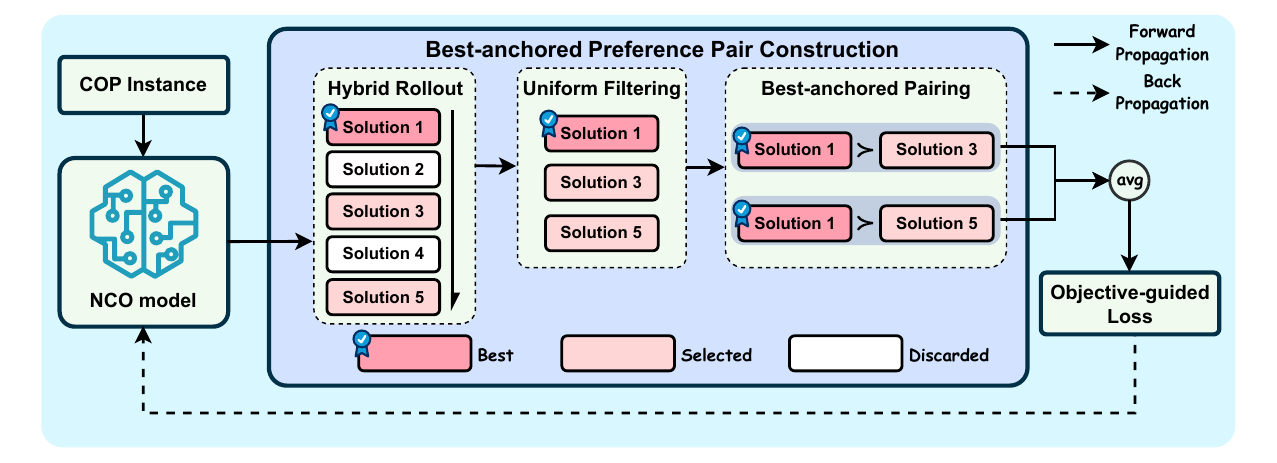}}
\caption{ The pipeline of best-anchored and objective-guided preference optimization (BOPO).}
\label{fg:bopo}
\vskip -0.2in
\end{figure*}

To improve sample efficiency in NCO training, we leverage multiple sampled solutions rather than focusing solely on the optimal one by introducing preference optimization \cite{dpo,simdpo}. To this end, we propose \textit{ Best-anchored and Objective-guided Preference Optimization (BOPO)}, which leverages the natural preference relation among solutions according to their objective values. Our approach builds upon two fundamental observations: (1) NCO models (typically generative models) can generate multiple distinct solutions for a given problem instance, and (2) the objective value of a COP solution can be computed with a low cost. As shown in Figure \ref{fg:bopo}, our { BOPO} comprises two essential components: constructing multiple preference pairs from sampled solutions and building a preference optimization loss. As a new training paradigm for neural combinatorial optimization, { BOPO} avoids expensive labels and formulation of Markov Decision Process, making it particularly well-suited for addressing various types of COPs.

In summary, our contributions are as follows: 
\begin{itemize}
   \item We propose { BOPO}, a novel training paradigm using preference optimization for neural combinatorial optimization, which enhances sample efficiency compared with the mainstream RL and recent SLL paradigms.
   \item As the first key component of { BOPO}, we design a { best-anchored} preference pair construction method for COPs to better explore and exploit solutions. 
   \item As the second key component, we tailor a novel { objective-guided} preference optimization loss that incorporates preferences quantified by objective values of COPs.
   \item Experimental results on three classic problems, namely the Job-shop Scheduling Problem (JSP), Traveling Salesman Problem (TSP), and Flexible Job-shop Scheduling Problem (FJSP), demonstrate that { BOPO} outperforms state-of-the-art methods.
\end{itemize}


\section{Related Works}
\label{sec:relatedworks}

\textbf{Supervised Learning (SL) for NCO.} SL methods utilize optimal solutions as labels to train neural models with cross-entropy loss for solving COPs, such as TSP \cite{pn,sul1} and JSP \cite{s26}. Data augmentation techniques have been exploited to enhance the performance of SL methods for routing problems \cite{sl1,tsp_sl}. Additionally, diffusion-based SL approaches \cite{sl2,sl3,sl4} learn to generate heatmaps for TSP. The primary limitation of SL, however, lies in the high computational cost of obtaining optimal solutions as labels, which restricts its practical applications.

\textbf{Reinforcement Learning (RL) \& Self-Labeling Learning (SLL) for NCO.} Label-free RL is currently the mainstream training paradigm in neural combinatorial optimization. Attention Model \cite{tsp_am}, which combines RL with Transformer \cite{transformer}, marks a milestone in solving routing problems. The policy optimization with multiple optima (POMO) \cite{tsp_pomo} introduces a shared baseline leveraging solution symmetry. Given its competitive performance and practicality, POMO has established itself as a prominent training algorithm for routing problems and has inspired numerous advancements \cite{gri23,cha23,dra23,che23,che23b,xia24,goh24,fan24,zho24,zho24b,bij24,zhe24,wan24,che25,che25b}. Additionally, RL has been widely adopted for scheduling problems, including JSP \cite{l2d,schedule_net,jsp_gnn,jsp_oneshot,s47,cl,rs} and FJSP \cite{fjsp0,fjsp2}. Different from the above RL-based constructive methods, RL is also employed in improvement methods for both routing \cite{iter2,iter1,iter3,iter4} and scheduling problems \cite{nls,jsp_iter}. Recently, SLL \cite{sil,slim} utilizes the local optimal solution during training as a pseudo-label to train an end-to-end model using cross-entropy loss, where self-labeling improvement method (SLIM) \cite{slim} has achieved state-of-the-art performance on JSP.

\textbf{Preference Optimization.} Preference optimization has been widely adopted to align large language models (LLMs) with human preferences. One of the most well-known techniques is reinforcement learning with human feedback (RLHF) \cite{rlhf,gpt}, which trains a reward model using ranking learning and then aligns LLM through RL. Direct preference optimization (DPO) \cite{dpo} offers an efficient alternative by skipping the reward model training phase and directly optimizing LLMs using preference pairs. Building upon DPO, subsequent studies explore comparing more samples \cite{d26,d72} and designing more concise loss functions \cite{d89,simdpo}. Among them, simple preference optimization (SimPO) \cite{simdpo} has gained popularity due to its simplicity and effectiveness. Recent advancements in preference optimization are summarized in a comprehensive survey \cite{dpo_survey}. Inspired by preference optimization, we propose { best-anchored and objective-guided preference optimization (BOPO)}. {  A concurrent work \cite{po4cop} also applies preference optimization to COPs. However, fundamentally different from \citet{po4cop}, BOPO develops a best-anchored preference pair construction method and a novel objective-guided preference optimization loss specially designed for COPs.}


\section{Preliminaries}
\label{sec:preliminaries}

\subsection{Neural Combinatorial Optimization (NCO)}
\label{sec:nco}

COP aims to find a solution $\bm y $ that minimizes (or maximizes) the objective function $g(\bm y)$. In the NCO domain, neural constructive methods sequentially construct a solution $\bm y$ in an end-to-end manner for a COP instance $\bm x$. Specifically, at step $t \in \{1, \cdots, |\bm y|\}$, a feasible action $y_t$ is selected based on the partial solution $\bm{y}_{<t}=(y_1,\cdots,y_{t-1})$ with constraints enforced through masking. A model with parameter $\bm\theta$ outputs the policy $\pi_{\bm\theta}(\bm y|\bm x)= \prod_{t=1}^{|\bm y|}\pi_{\bm\theta}(y_t|\bm{y}_{<t},\bm x)$ of solution $\bm y$. Solutions can be obtained via multiple search strategies based on policy $\pi_{\bm\theta}(\bm y|\bm x)$, including greedy and sampling rollouts.

Typical training paradigms include SL, RL, and SLL. SL utilizes the (near-) optimal solution $\bm{y}^*$ as a label to train model $\bm\theta$ using cross-entropy loss $\mathcal L(\pi_{\bm\theta}|\bm{y}^*,\bm x)=-\log \pi_{\bm\theta}(\bm{y}^*|\bm x)$.
In RL, the REINFORCE loss \cite{reinforce} $\mathcal L(\pi_{\bm\theta}|\bm y,\bm x)=-(g(\bm y)-b(\bm x))\log\pi_{\bm\theta}(\bm y|\bm x)$ with a baseline $b(\bm x)$ is commonly used for routing problems, 
while the proximal policy optimization (PPO) \cite{ppo} loss $\mathcal L(\pi_{\bm\theta}|\bm y,\bm x)=-\sum_t\min\{r_t\hat A_t,\text{clip}(r_t,1-\epsilon,1+\epsilon)\hat A_t\}$ is predominantly used for scheduling problems, where $r_t=\frac{\pi_{\bm\theta}(y_t|\bm{y}_{<t},\bm x)}{\pi_{{\bm\theta}^\prime}(y_t|\bm{y}_{<t},\bm x)}$ denotes the ratio of current policy $\bm\theta$ and old policy ${\bm\theta}^\prime$, $\hat A_t$ represents the advantage estimate, and $\epsilon$ is a hyperparameter. SLL, a recent paradigm, selects the best among solutions sampled from current policy $\bm\theta$ as a pseudo-label and applies cross-entropy loss.

\subsection{Solution Construction for Classic COPs}
\label{sec:cop_example}

The job-shop scheduling problem (JSP) entails allocating a set of $n$ jobs across $m$ machines with shape $(n\times m)$, wherein each job must be performed on the machines in a predefined sequence. The instance $\bm x$ comprises the processing time of operations and the corresponding required machines. Action $y_t$ is defined as an operation that assigns a job to the earliest available time slot on the corresponding machine and updates the job's progress. The goal is to determine the job processing order on each machine to minimize the maximum completion time, known as the \emph{makespan}. This construction involves $|\bm y|=nm$ steps.

For the traveling salesman problem (TSP), the instance $\bm x$ is composed of $n$ nodes with 2-dimensional coordinates. The objective is to find a tour that passes through all nodes with minimal total distance. To construct a TSP solution, an unvisited node $y_t$ at step $t$ is selected to be added to the current partial tour. This process requires $|\bm y|=n$ steps.

The flexible job-shop scheduling problem (FJSP) extends JSP by considering that each operation can be processed on multiple candidate machines with shape $(n\times m \times k)$, where $k$ denotes the maximum number of operations in all jobs. To construct a FJSP solution, action $y_t$ at step $t$ represents a joint selection of an operation and one of its available machines. This construction requires $|\bm y|=nk$ steps.

The definitions of above COPs and details of their features are provided in \cref{ap:problem}.


\section{Methodology}
\label{sec:methodology}

\subsection{ Best-anchored and Objective-guided Preference Optimization (BOPO)}
\label{sec:bopo}

To improve sample efficiency, we propose a novel training paradigm, namely { BOPO}. Distinct from the RL and SLL training paradigms, { BOPO} exploits the preference relations among generated solutions according to their objective values. Specifically, for a COP with the minimization objective, the \textit{explicit preference} $f^*(\bm y,\bm x)$ for instance $\bm x$ and solution $\bm y$ is defined as the negative of the objective function: 
\begin{equation}
\label{ex_pre}
f^*(\bm y,\bm x)=-g(\bm y).
\end{equation}
A \textit{preference pair}, denoted as a triplet $(\bm x,\bm{y}_w,\bm{y}_l)$, consists of an instance $\bm x$ and two solutions $\bm{y}_w$ and $\bm{y}_l$ satisfying $\bm{y}_w \succ \bm{y}_l\triangleq f^*(\bm{y}_w,\bm x)>f^*(\bm{y}_l,\bm x)$.

{ BOPO} employs a preference optimization loss based on such preference pairs to train the neural model parameterized as $\bm \theta$. As the two critical components of { BOPO}, we develop a { best-anchored} preference pair construction method and derive a novel { objective-guided preference optimization} loss function specialized for COPs.

\subsection{{ Best-anchored} Preference Pair Construction}
\label{sec:pairs}
The construction of preference pairs consists of three steps: (1) \textbf{Hybrid Rollout} generates diverse solutions via sampling rollout and a high-quality one via greedy rollout. (2) \textbf{Uniform Filtering} selects representative ones from the obtained solutions for efficient pairing. (3) \textbf{Best-anchored Pairing} constructs preference pairs to enhance model learning.

\textbf{Hybrid Rollout.} Both diverse and high-quality solutions play vital roles in model learning. Sampling from policy $\pi_{\bm\theta}(\bm y,\bm x)$ generates diverse solutions, occasionally surpassing the greedy rollout solution. However, most sampled solutions are inferior to the greedy one. To leverage their complementary strengths, we propose a hybrid rollout strategy combining both approaches. This strategy generates $B$ solutions, including $B-1$ from sampling and one from greedy rollout. It ensures coverage of both exploratory and exploitative solutions.

\textbf{Uniform Filtering.} Constructing preference pairs using all $B$ solutions would produce a combination of $\binom{B}{2}$ pairs, resulting in high computational cost and many low-quality pairs. Instead, we employ uniform filtering to select solutions to maximize representational diversity. Specifically, we select $K$ solutions $\mathcal C=\{\bm{y}_1\succ\cdots\succ\bm{y}_K\}$ uniformly from $B$ sorted solutions $\mathcal S= \{\bm{y}'_1 \succ \cdots \succ \bm{y}'_B \}$, i.e., $\bm{y}_k=\bm{y}'_{\lfloor B/K \rfloor\cdot(k-1)+1}$, $\forall k \in \{1, \cdots, K\}$. This avoids overfitting to clusters of similar solutions.

\textbf{Best-anchored Pairing.} Since a COP solely focuses on finding the optimal solution, we anchor pairs to the best solution to prioritize learning from high-quality examples. 
For $K$ solutions $\{\bm{y}_1\succ\cdots\succ\bm{y}_K\}$, we create $K-1$ preference pairs, each combining the best solution with a suboptimal one, i.e., $\mathcal P= \{(\bm{x}, \bm{y}_1, \bm{y}_k) | k\in\{2,\cdots,K\}\}$. This design encourages learning from the optimal solution while discouraging learning from various suboptimal ones, being more efficient than using all $\binom{K}{2}$ possible pairs.

\subsection{{ Objective-guided Preference Optimization} Loss}
\label{sec:bopo_loss}

After obtaining preference pairs, we formulate the loss function of { BOPO} by incorporating a preference-based scaling factor derived from the objective values of COP solutions.

\textbf{{ Objective-guided Preference Optimization} Loss.} For policy $\pi_{\bm\theta}(\bm y,\bm x)$ used to construct solution $\bm y$, its \textit{implicit preference} $f_{\bm\theta}(\bm y,\bm x)$ is defined as the average log-likelihood:
\begin{equation}
\label{im_pre}
f_{\bm\theta}(\bm y,\bm x)
=\frac{1}{|\bm y|}\log \pi_{\bm\theta}(\bm y|\bm x) 
=\frac{1}{|\bm y|}\sum_{t=1}^{|\bm y|}\log \pi_{\bm\theta}(y_t|\bm{y}_{<t},\bm x).
\end{equation}
For a preference pair $(\bm x,\bm{y}_w,\bm{y}_l)$, the \textit{preference distribution} $p_{\bm\theta}(\bm{y}_w \succ \bm{y}_l|\bm x)$ is modeled using the Bradley-Terry ranking objective \cite{bt_model} and implicit preferences:
\begin{equation}
\label{p_pre}
\begin{aligned}
p_{\bm\theta}(\bm{y}_w & \succ \bm{y}_l|\bm x)= \\
& \sigma (\beta(\bm x,\bm{y}_w,\bm{y}_l) (f_{\bm\theta}(\bm{y}_w,\bm x)-f_{\bm\theta}(\bm{y}_l,\bm x))),
\end{aligned}
\end{equation}
where $\sigma(\cdot)$ is the sigmoid function. $\beta(\bm x,\bm{y}_w,\bm{y}_l)=f^*(\bm{y}_l,\bm x)/f^*(\bm{y}_w,\bm x)$ is a preference-based adaptive scaling factor derived from explicit preferences, which acts as a \emph{natural curriculum}. For different pairs with the same best solution $\bm{y}_w$ but different suboptimal solutions $\bm{y}_l$, their preference differences should vary according to explicit preferences. Therefore, it is wise to introduce a scaling factor to adjust the difference in the preference distribution.

By maximizing the log-likelihood of $p_{\bm\theta}(\bm{y}_w \succ \bm{y}_l|\bm x)$, the model is encouraged to assign higher probabilities to preferred solutions $\bm{y}_w$ compared with less preferred solutions $\bm{y}_l$. From \cref{ex_pre,im_pre,p_pre}, we can derive the { BOPO} loss function:

{\small
\begin{equation}\label{loss}\begin{aligned}
& \mathcal L_{BOPO}(\pi_{\bm\theta},\bm x,\bm{y}_w,\bm{y}_l) =  \\
& 
-\log \sigma\left( \underbrace{\frac{g(\bm{y}_l)}{g(\bm{y}_w)}}_{\text{Adaptive Scaling}} \left( \underbrace{\frac{\log \pi_\theta(\bm{y}_w|\bm{x})}{|\bm{y}_w|}  - \frac{\log \pi_\theta(\bm{y}_l|\bm{x})}{|\bm{y}_l|} }_{\text{Average Log-likelihood Difference}} \right) \right).
\end{aligned}\end{equation}
}

The { BOPO} training algorithm is presented in \cref{alg:bopo}.

\begin{algorithm}[tb]
    \caption{{ BOPO} Training}
    \label{alg:bopo}
\begin{algorithmic}[1]
\STATE {\bfseries Input:} Dataset $\mathcal X$, number of epochs $E$, number of training steps $T$, batch size $D$, number of obtained solutions $B$, number of filtered solutions $K$, and learning rate $\eta$ 
\STATE Initialize model parameter $\bm\theta$
\FOR{$epoch=1$ {\bfseries to} $E$}
\FOR{$step=1$ {\bfseries to} $T$}
    \STATE $\bm{x}_i \gets$ \textsc{SampleInstance}($\mathcal X$) $\forall i\in\{1,\dots,D\}$
    \STATE $\mathcal S_i \gets $ \textsc{HybridRollout}($\bm{x}_i,B$) $\forall i\in\{1,\dots,D\}$
    \STATE \resizebox{\linewidth}{!}{$\mathcal C_i \gets $ \textsc{UniformFiltering}($\mathcal S_i,K$) $\forall i\in\{1,\dots,D\}$}
    \STATE \resizebox{\linewidth}{!}{$\mathcal P_i \gets $ \textsc{Best-anchoredPairing}($\mathcal C_i$) $\forall i\in\{1,\dots,D\}$}
    \STATE \resizebox{\linewidth}{!}{Compute $\mathcal{L}_{BOPO} (\pi_{\bm\theta}, \bm{x}, \bm{y}_w, \bm{y}_l)$ using \cref{loss}}
    \STATE \resizebox{\linewidth}{!}{$\mathcal{L}(\bm\theta) \gets \frac{1}{D} \sum_{i=1}^{D} \frac{1}{|\mathcal P_i|}\sum_{(\bm{x},\bm{y}_w,\bm{y}_l) \in \mathcal P_i}{ \mathcal{L}_{BOPO}} (\pi_{\bm\theta}, \bm{x}, \bm{y}_w, \bm{y}_l) $}
    \STATE $\bm\theta \gets \text{Adam}( \bm\theta ,  \nabla_{\bm\theta}\mathcal{L}(\bm\theta), \eta)$
    \ENDFOR
\ENDFOR
\end{algorithmic}
\end{algorithm}

\textbf{Comparison with Other Losses.} Our { BOPO} loss differs from existing preference optimization losses in several key aspects. Compared with RLHF \cite{rlhf}, it eliminates the need to train an additional reward model. Compared with DPO \cite{dpo}, it avoids using a reference model, reducing computational costs. Compared with SimPO \cite{simdpo}, it incorporates a { objective-guided} scaling factor without requiring extra hyperparameters, avoiding labor-intensive hyperparameter tuning. Detailed analyses are provided in \cref{ap:other_methods}.

{ 
\textbf{Comparison with RL.} By leveraging preference learning between the best solution and diverse inferior ones, BOPO guides the model toward promising decision trajectories and discern suboptimal choices, demonstrating its advantage against RL methods. Meanwhile, recognizing that precise rewards (i.e., objective values) in RL are crucial to combinatorial optimization, BOPO introduces an objective-guided scaling factor beyond the standard preference optimization loss function to better distinguish preference differences.
}

\subsection{Characteristics of { BOPO}}

In summary, our { BOPO} has the following characteristics. (1) \textbf{Novel training paradigm}: { BOPO} introduces preference optimization to neural combinatorial optimization as a new training paradigm, featuring two effective problem-awareness components: { best-anchored} preference pair construction and { objective-guided preference optimization} loss. (2) \textbf{Architecture agnostic}: { BOPO} is compatible with various models for different problems, achieving high sample efficiency without expensive labels while inheriting the fast inference advantage of neural models. (3) \textbf{Outstanding performance}: Compared with existing methods, { BOPO} surpasses state-of-the-art results on classic COPs, including JSP, TSP, and FJSP.


\section{Experimental Results}
\label{sec:results}

To evaluate the performance of { BOPO}, we compare it with state-of-the-art NCO methods and strong traditional solvers on typical COP benchmarks with various problem shapes and distributions. Performance evaluation is based on the gap metric $\frac{g(\bm y)-g(\bm{y}^*)}{g(\bm{y}^*)}\times100 \%$ between the obtained solution $\bm y$ and known optimal solution $\bm{y}^*$, where a lower gap indicates better performance. The best results are highlighted in \textbf{bold}. We also report total solving time for each instance group. Experiments were conducted on a Linux system with an NVIDIA TITAN Xp GPU and an Intel(R) Xeon(R) E5-2680 CPU. {  Our implementation of BOPO using PyTorch and trained models for each problem are available.\footnote{ \href{https://github.com/L-Z-7/BOPO}{https://github.com/L-Z-7/BOPO}}}

\begin{table*}[thb!]
    \centering
    \caption{Average gaps (\%) of evaluated methods on JSP benchmarks. ``-" indicates unavailable results from the corresponding paper.}
    \label{tb:jsp_main}
\addtolength{\tabcolsep}{-3pt}
\vskip 0.15in
\resizebox{\textwidth}{!}{
\begin{tabular}{rl||cc|cccc|ccc|ccc|cc|c|ccccc|ccc}  \toprule \midrule
&  & \multicolumn{6}{c|}{Non-constructive} & \multicolumn{9}{c|}{Greedy Constructive}  & \multicolumn{8}{c}{Sampling Constructive} \\
&  & \multicolumn{2}{c}{Exact Solver} & \multicolumn{4}{c|}{RL-based Improvement}  & \multicolumn{3}{c}{Traditional PDR} & \multicolumn{3}{c}{RL} & \multicolumn{2}{c}{SLL} & { BOPO} & \multicolumn{5}{c}{$B'=128$} & \multicolumn{3}{c}{$B'=512$} \\ \cmidrule(lr){3-4}\cmidrule(lr){5-8}\cmidrule(lr){9-11}\cmidrule(lr){12-14}\cmidrule(lr){15-16}\cmidrule(lr){17-17}\cmidrule(lr){18-22}\cmidrule(lr){23-25}
& Shape  & Gurobi  & OR-Tools & L2S$_{500}$ & NLS$_A$ & TGA$_{500}$ & L2S$_{5k}$ & SPT    & MOR    & MWR    & L2D   & SchN & CL & SLIM$_\text{MGL}$  & SLIM  & { BOPO} & L2D  & CL  & SLIM$_\text{MGL}$  & SLIM  & { BOPO}  & SLIM$_\text{MGL}$  & SLIM  & { BOPO}  \\ \midrule \toprule
\multirow{9}{*}{\rotatebox{90}{TA}} & 15$\times$15  & 0.1  & 0.1 & 9.3  & 7.7  & 8.0  & 6.2  & 25.8 & 20.5 & 19.2 & 26.0 & 15.3 & 14.3 & 13.1 & 13.8          & 13.6          & 17.1 & 9.0  & 8.8          & 7.2           & 7.1          & 7.2          & 6.5          & \textbf{6.3}  \\
& 20$\times$15  & 3.2  & 0.2 & 11.6 & 12.2 & 9.9  & 8.3  & 32.9 & 23.6 & 23.4 & 30.0 & 19.4 & 16.5 & 16.1          & 15            & 14.3 & 23.7 & 10.6 & 11.0         & 9.3           & 9.0          & 10.4         & 8.8          & \textbf{8.3}  \\
& 20$\times$20  & 2.9  & 0.7 & 12.4 & 11.5 & 10.0 & 9.0  & 27.8 & 21.7 & 21.8 & 31.6 & 17.2 & 17.3 & 15.3          & 15.2          & 15.1 & 22.6 & 10.9 & 11.1         & 10.0          & 9.8          & 10.0         & \textbf{9.0} & 9.1           \\
& 30$\times$15$^*$  & 10.7 & 2.1 & 14.7 & 14.1 & 13.3 & 9.0  & 35.1 & 22.7 & 23.7 & 33.0 & 19.1 & 18.5 & 17.7          & 17.1          & 16.6 & 24.4 & 14.0 & 14.0         & 11.0          & 11.0         & 12.2         & 10.6         & \textbf{10.3} \\
& 30$\times$20$^*$  & 13.2 & 2.8 & 17.5 & 16.4 & 16.4 & 12.6 & 34.4 & 24.9 & 25.2 & 33.6 & 23.7 & 21.5 & 19.3          & 18.5          & 17.1 & 28.4 & 16.1 & 16.3         & 13.4          & 13.3         & 14.9         & 12.7         & \textbf{12.2} \\
& 50$\times$15$^*$  & 12.2 & 3.0 & 11.0 & 11.0 & 9.6  & 4.6  & 24.1 & 17.3 & 16.8 & 22.4 & 13.9 & 12.2 & 13.4          & 10.1          & 9.8  & 17.1 & 9.3  & 9.2          & 5.5           & 5.8          & 8.2          & \textbf{4.9} & \textbf{4.9}  \\
& 50$\times$20$^*$  & 13.6 & 2.8 & 13.0 & 11.2 & 11.9 & 6.5  & 25.6 & 17.7 & 17.9 & 26.5 & 13.5 & 13.2 & 14.0          & 11.6 & 11.8          & 20.4 & 9.9  & 10.6         & 8.4           & 8.0          & 9.8          & 7.6          & \textbf{7.4}  \\
& 100$\times$20$^*$ & 11.0 & 3.9 & 7.9  & 5.9  & 6.4  & 3.0  & 14.4 & 9.2  & 8.3  & 13.6 & 6.7  & 5.9  & 7.4           & 5.8           & 4.9  & 13.3 & 4.0  & 4.8          & 2.3           & 1.8          & 4.4          & 2.1          & \textbf{1.4}  \\ \midrule
& Avg    & 8.4  & 2.0 & 12.2 & 11.3 & 10.7 & 7.4  & 27.5 & 19.7 & 19.5 & 27.1 & 16.1 & 14.9 & 14.5          & 13.4          & 12.9 & 20.8 & 10.4 & 10.7         & 8.4           & 8.2          & 9.6          & 7.8          & \textbf{7.5}  \\ \midrule \toprule
\multirow{8}{*}{\rotatebox{90}{LA}} & 10$\times$5$^*$   & 0.0  & 0.0 & 2.1  & -    & 2.1  & 1.8  & 14.8 & 16.0 & 16.0 & 14.3 & 12.1 & -    & 8.6           & 9.3           & 6.0  & 8.8  & -    & 3.7          & 1.9           & 2.7          & 2.5          & \textbf{1.1} & 2.1           \\
& 10$\times$10  & 0.0  & 0.0 & 4.4  & -    & 1.8  & 0.9  & 15.7 & 18.1 & 12.2 & 23.7 & 11.9 & -    & 9.1           & 8.9           & 8.2  & 10.4 & -    & 3.5          & 3.1           & 2.3          & 2.4          & 2.5          & \textbf{2.1}  \\
& 15$\times$5$^*$   & 0.0  & 0.0 & 0.0  & -    & 0.0  & 0.0  & 14.9 & 3.9  & 5.5  & 7.8  & 2.7  & -    & 1.5           & 2.6           & 1.1  & 2.8  & -    & \textbf{0.0} & \textbf{0.0}  & \textbf{0.0} & \textbf{0.0} & \textbf{0.0} & \textbf{0.0}  \\
& 15$\times$10  & 0.0  & 0.0 & 6.4  & -    & 3.6  & 3.4  & 28.7 & 23.7 & 17.8 & 27.2 & 14.6 & -    & 11.7          & 11.6          & 11.0 & 16.2 & -    & 6.3          & 5.2           & 5.8          & 5.6          & 5.0          & \textbf{4.9}  \\
& 15$\times$15  & 0.0  & 0.0 & 7.3  & -    & 5.5  & 5.9  & 24.6 & 18.1 & 18.2 & 27.1 & 16.1 & -    & 13.5          & 13.6          & 12.2 & 17.4 & -    & 7.1          & 6.8           & 6.5          & 6.7          & 5.6          & \textbf{4.9}  \\
& 20$\times$5$^*$   & 0.0  & 0.0 & 0.0  & -    & 0.0  & 0.0  & 13.7 & 3.8  & 5.2  & 6.3  & 3.6  & -    & 1.5           & 2.1           & 0.4  & 3.1  & -    & 0.5          & \textbf{0.0}  & \textbf{0.0} & \textbf{0.0} & \textbf{0.0} & \textbf{0.0}  \\
& 20$\times$10  & 0.0  & 0.0 & 7.0  & -    & 5.0  & 2.6  & 33.4 & 20.9 & 17.2 & 24.6 & 15.7 & -    & 14.3          & 12.1 & 12.2          & 18.3 & -    & 7.9          & 6.9           & 5.9          & 7.1          & 5.6          & \textbf{4.6}  \\
& 30$\times$10$^*$  & 0.0  & 0.0 & 0.2  & -    & 0.0  & 0.0  & 13.9 & 6.5  & 8.6  & 8.4  & 3.1  & -    & 3.1           & 2    & 2.4           & 6.8  & -    & 0.3          & \textbf{0.0}  & \textbf{0.0} & 0.1          & \textbf{0.0} & \textbf{0.0}  \\ \midrule
& Avg    & 0.0  & 0.0 & 3.4  & -    & 2.3  & 1.8  & 20.0 & 13.9 & 12.6 & 17.4 & 10.0 & -    & 7.9           & 7.8           & 6.7  & 10.6 & -    & 3.7          & 3.0           & 2.9          & 3.0          & 2.5          & \textbf{2.3} \\
\midrule \toprule
\multirow{9}{*}{\rotatebox{90}{DMU}} & 20$\times$15 & 5.3  & 1.8 & - & - & - & - & 28.0 & 30.9 & 28.8 & 39.0 & - & - & 17.0 & 18   & 17.5 & 29.3 & 19.4 & 13.7 & 12.0 & 11.2 & 12.7 & 11.3          & \textbf{10.4} \\
                     & 20$\times$20 & 4.7  & 1.9 & - & - & - & - & 31.3 & 27.4 & 27.3 & 37.7 & - & - & 22.6 & 19.4 & 20.3 & 27.1 & 16.0 & 15.3 & 13.5 & 12.7 & 14.1 & 12.3          & \textbf{11.8} \\
                     & 30$\times$15$^*$ & 14.2 & 2.5 & - & - & - & - & 31.5 & 37.4 & 32.3 & 42.0 & - & - & 24.1 & 21.8 & 19.1 & 34.0 & 16.5 & 18.4 & 14.4 & 13.9 & 17.5 & 14.0          & \textbf{12.9} \\
                     & 30$\times$20$^*$ & 16.7 & 4.4 & - & - & - & - & 34.4 & 34.7 & 31.4 & 39.7 & - & - & 25.6 & 25.7 & 25.6 & 33.6 & 20.2 & 19.0 & 17.1 & 16.5 & 17.8 & 15.8          & \textbf{15.5} \\
                     & 40$\times$15$^*$ & 16.3 & 4.1 & - & - & - & - & 24.0 & 36.7 & 27.5 & 35.6 & - & - & 20.1 & 17.5 & 15.9 & 31.5 & 17.6 & 15.8 & 11.7 & 11.4 & 15.3 & \textbf{10.9} & \textbf{10.9} \\
                     & 40$\times$20$^*$ & 22.5 & 4.6 & - & - & - & - & 37.2 & 37.1 & 32.9 & 39.6 & - & - & 23.5 & 22.2 & 22.3 & 35.8 & 25.6 & 19.8 & 16.0 & 16.7 & 19.0 & \textbf{14.8} & 15.9          \\
                     & 50$\times$15$^*$ & 14.9 & 3.8 & - & - & - & - & 24.8 & 35.5 & 28.0 & 36.5 & - & - & 18.2 & 15.7 & 14.5 & 32.7 & 21.7 & 15.6 & 11.2 & 11.2 & 15.3 & 10.6          & \textbf{10.4} \\
                     & 50$\times$20$^*$ & 22.5 & 4.8 & - & - & - & - & 30.1 & 37.0 & 30.8 & 39.5 & - & - & 25.8 & 22.4 & 25.2 & 36.1 & 15.2 & 20.8 & 15.8 & 16.5 & 20.0 & \textbf{15.0} & 15.5          \\ \midrule
                     & Avg   & 14.6 & 3.5 & - & - & - & - & 30.2 & 34.6 & 29.9 & 38.7 & - & - & 22.1 & 20.3 & 20.0 & 32.5 & 19.0 & 17.3 & 14.0 & 13.8 & 16.5 & 13.1          & \textbf{12.9}\\
\midrule \bottomrule
\end{tabular}}
\vskip -0.1in
\end{table*}

\subsection{Job-shop Scheduling Problem}
\label{sec:res_jsp} 

\textbf{Neural Model.}
Each JSP instance is represented as a disjunctive graph, a standard representation for scheduling problems. For details of disjunctive graphs, see \cref{ap:problem}. We employ a neural model, named MGL, that combines a multi-layer graph attention network (GAT) \cite{gat} encoder for computing node embeddings with a long short-term memory (LSTM) based \cite{lstm} context-attention decoder for predicting action probabilities using both embedding and context features. The complete architecture is detailed in \cref{ap:nn}.

\textbf{Training \& Test.} 
For evaluation, we use three standard JSP benchmarks: Taillard's (TA) \cite{benchmark_ta}, Lawrence's (LA) \cite{benchmark_la}, and Demirkol's (DMU) \cite{benchmark_dmu}. Each benchmark contains 8 different shapes with 10 instances per shape, except LA which has 5 instances per shape. We generate a training dataset of 30000 instances following SLIM \cite{slim}, consisting of 6 shapes $(n\times m)$ in $\{10\times10,15\times10,15\times15,20\times10,20\times15,20\times20\}$ with 5000 instances per shape. During training, we generate additional 100 different instances per shape from the same shape set for validation. We employ the Adam optimizer \cite{adam} with learning rate $\eta=0.0002$ and train the neural model for 20 epochs. We set the solution number of hybrid rollout $B=256$, the number of filtered solutions $K=16$, batch size of $D=1$. During inference, we adopt both greedy rollout and sampling rollout with $B'$ solutions.

\textbf{Baselines.}
We compare { BOPO} with two categories of approaches: non-constructive methods and constructive methods. (1) Non-constructive methods, which require extensive search time, include both exact solvers and state-of-the-art neural improvement methods. We employ two exact solvers: \textbf{Gurobi} and Google \textbf{OR Tools}, both with a time limit of 3600 seconds. We also include four RL-based improvement methods: \textbf{NLS}$_A$ \cite{nls}, \textbf{L2S} with 500 (L2S${500}$) and 5000 (L2S${5k}$) solutions \cite{l2s}, and \textbf{TGA}${500}$ \cite{jsp_iter} with 500 solutions. (2) Constructive methods comprise widely used traditional constructive heuristics and state-of-the-art neural constructive methods, where neural methods adopt both greedy rollout and sampling rollout with $B'$ solutions. For traditional constructive heuristics, we consider three representative traditional Priority Dispatching Rules (PDRs) \cite{pdr1}: \textbf{shortest processing time} (SPT), \textbf{most operations remaining} (MOR), and \textbf{most work remaining} (MWR). The neural constructive baselines include three RL-based methods: \textbf{L2D} \cite{l2d} and \textbf{SchN} \cite{schedule_net}, which utilize PPO with different modeling approaches, and \textbf{CL} \cite{cl}, which incorporates curriculum learning. We also include two state-of-the-art SLL-based baselines, \textbf{SLIM} \cite{slim} and \textbf{SLIM$_\text{MGL}$}, where SLIM$_\text{MGL}$ uses our MGL model with SLIM's training algorithm. For a fair comparison, we set its batch size to 16, matching { BOPO}'s setting.

\textbf{Results on JSP Benchmarks.} 
Comparative results on the TA and LA benchmarks are presented in \cref{tb:jsp_main}. Our method achieves the lowest average optimality gap among all constructive methods on all benchmarks, with sampling rollout further enhancing its performance through the exploration of more solutions. Notably, { BOPO} even surpasses RL-based improvement methods, with the exception of L2S$_{5k}$, where { BOPO} achieves a comparable gap (7.5\% vs. 7.4\% on TA) while requiring significantly less computational time (4.8m vs. 4h on TA). Detailed runtime analysis is provided in \cref{ap:time}. Compared with SLIM, the current state-of-the-art SLL-based constructive method, { BOPO}, which employs the efficient MGL model, achieves both reduced parameter count and computational overhead (detailed in \cref{ap:nn}). More significantly, when evaluated against SLIM$_\text{MGL}$ using the identical model, the performance disparity increases markedly across all scenarios, underscoring the fundamental advantage of the proposed training paradigm over the SLL counterpart. Furthermore, { BOPO} exhibits superior generalization performance on out-of-distribution problem shapes (marked by $^*$).

\subsection{Traveling Salesman Problem}
\label{sec:res_tsp}

\textbf{Neural Model.}
{ BOPO adopts} the same model as the typical POMO \cite{tsp_pomo}, comprising an encoder with 6 Transformer layers and a decoder with a multi-head attention layer. 
{ To further demonstrate the universality of BOPO, it also employs the state-of-the-art INViT \cite{fan24} model.}

\textbf{Training \& Test.} 
Following the NCO literature, we evaluate the proposed method on randomly generated instances with $n=20/50/100$ (denoted as TSP20/50/100), using models trained on corresponding problem shapes. Training instances are generated randomly from an uniform distribution. Additionally, we assess { BOPO}'s generalization capability on the out-of-distribution TSPLIB benchmark \cite{tsplib}, {  and randomly generated instances with four different distributions, including uniform, cluster, explosion, and implosion distributions}. We adopt the same hyperparameter configuration as POMO {  and INViT, respectively}. During training, the number of filtered solutions is set to $K=8$. For TSP20 and TSP50, we set the hybrid rollout solution number $B=128$ and batch size $D=64$. For TSP100, we set $B=256,K=16$ and $D=48$ due to the memory limit. {  Specifically, we adopt BOPO to train INViT-2V, but due to memory constraints, we set $D=1,B=64$ (half of batch size in INViT), with $K=8$ and double epochs.} For inference, we adopt greedy rollout with multiple start nodes and $\times$8 instance augmentation (denoted as aug.), consistent with POMO. 

\textbf{Baselines.}
We assess { BOPO} against both traditional methods and neural constructive methods. The traditional methods include exact solver Concorde and Gurobi, and LKH3, a powerful problem-specific heuristic. The neural constructive methods include POMO \cite{tsp_pomo}, a widely-adopted RL-based backbone for advanced methods; {  DABL \cite{tsp_sl}, a state-of-the-art SL-based method with data augmentation for routing problems; INViT \cite{fan24}, a state-of-the-art RL-based method;} and SLIM \cite{slim}, a state-of-the-art SLL-based method {  applied to the POMO model.} 

\begin{figure*}[ht!]
\vskip 0.2in
    \centering
    \begin{subfigure}[b]{0.24\textwidth} 
        \includegraphics[width=\textwidth]{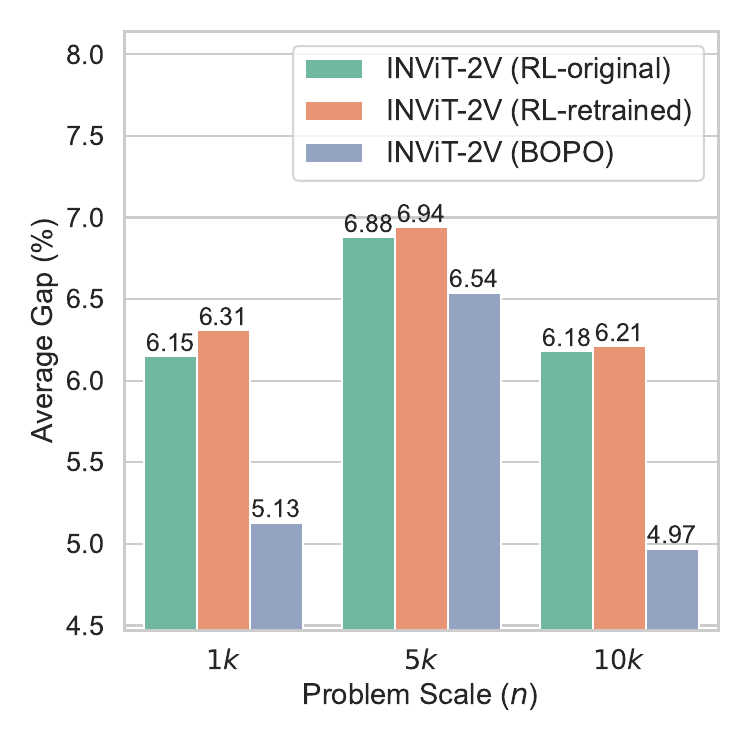}
        \caption{}
        \label{fg:invit-u}
    \end{subfigure}
    \begin{subfigure}[b]{0.24\textwidth}
        \includegraphics[width=\textwidth]{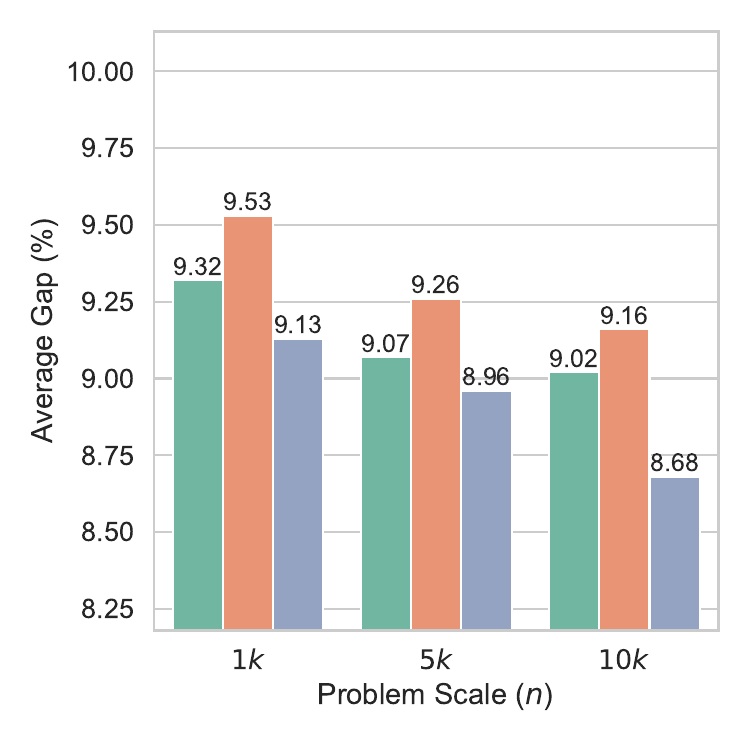} 
        \caption{}
        \label{fg:invit-c}
    \end{subfigure}
    \begin{subfigure}[b]{0.24\textwidth}
        \includegraphics[width=\textwidth]{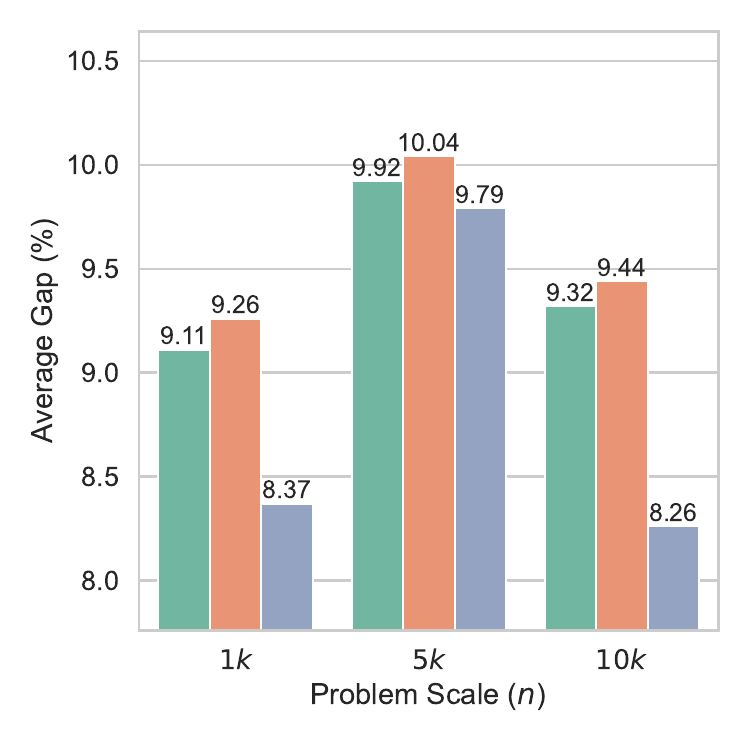}
        \caption{}
        \label{fg:invit-e}
    \end{subfigure}
    \begin{subfigure}[b]{0.24\textwidth}
        \includegraphics[width=\textwidth]{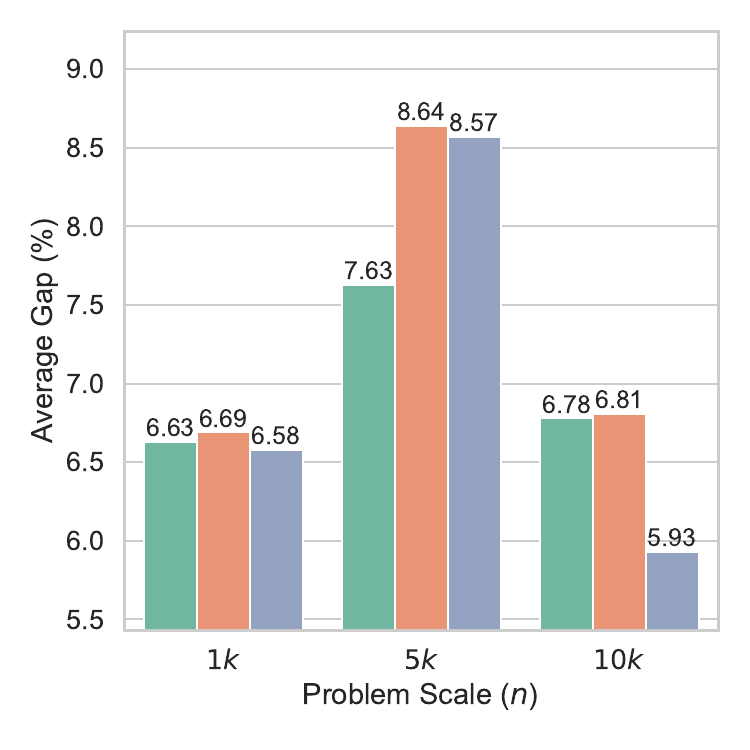}
        \caption{}
        \label{fg:invit-i}
    \end{subfigure}
    \caption{{ Performance on TSP with different problem scales and distributions: (a) uniform, (b) cluster, (c) explosion, and (d) implosion. RL-original denotes the results reported in the INViT paper; RL-retrained denotes the results retrained with the same batch size as BOPO.}}
    \label{fg:invit}
\vskip -0.2in
\end{figure*}

\textbf{Results on TSP}. 
{ Results based on POMO are presented in \cref{tb:tsp_main} and \cref{tb:tsp_gen}.} As shown in \cref{tb:tsp_main}, { BOPO} outperforms other neural baselines (except DABL on TSP50) while delivering competitive solutions against traditional solvers, despite the latter requiring substantially more computational time. { Compared with DABL which requires expensive labeled optimal solutions for SL, BOPO achieves comparable performance, demonstrating its superiority.} Notably, our BOPO achieves superior performance over POMO on TSP100, even with fewer training epochs (700 vs. 2000). Generalization results on TSPLIB presented in \cref{tb:tsp_gen} demonstrate { BOPO}'s significant advantages over other neural baselines when generalizing to out-of-distribution instances. { Moreover, results based on INViT's generalization to other distributions are shown in \cref{fg:invit}, showing consistent superiority when implemented on different models.}

\begin{table}[tb!]
\centering
    \caption{Results on 1000 uniformly generated TSP instances.}
    \label{tb:tsp_main}
\vskip 0.15in
\resizebox{\linewidth}{!}{
\addtolength{\tabcolsep}{-3pt}
\begin{tabular}{l||ccc|ccc|ccc}
\toprule
\midrule
\multirow{2}{*}{Method} & \multicolumn{3}{c|}{TSP20} & \multicolumn{3}{c|}{TSP50} & \multicolumn{3}{c}{TSP100} \\
 & Obj.$\downarrow$    & Gap$\downarrow$    & Time$\downarrow$  & Obj.$\downarrow$    & Gap$\downarrow$    & Time$\downarrow$ & Obj.$\downarrow$    & Gap$\downarrow$    & Time$\downarrow$ \\ \midrule \toprule
{ Concorde}    & 3.83  & 0.00  & 5m    & 5.69  & 0.00  & 13m  & 7.75  & 0.00  & 1h   \\
Gurobi      & 3.83  & 0.00  & 7s    & 5.69  & 0.00  & 2m   & 7.75  & 0.00  & 17m  \\
LKH3        & 3.83  & 0.00  & 42s   & 5.69  & 0.00  & 6m   & 7.75  & 0.00  & 25m  \\ \midrule
POMO        & 3.83  & 0.04  & 3.3s  & 5.70  & 0.21  & 6.4s & 7.80  & 0.46  & 11.4s \\
{ DABL}        & 3.83  & 0.01  & 3.3s  & 5.69  & 0.04  & 6.4s & 7.77  & 0.29  & 11.4s \\
SLIM        & 3.85  & 0.22  & 3.3s  & 5.78  & 1.51  & 6.4s & 8.18  & 5.51  & 11.4s \\
{ BOPO}        & 3.83  & 0.02  & 3.3s  & 5.70  & 0.14  & 6.4s & 7.78  & 0.37  & 11.4s \\ \midrule
POMO (aug.) & 3.83  & 0.00  & 3.6s & 5.69  & 0.03  & 6.6s  & 7.77  & 0.14  & 18.1s \\
{ DABL (aug.)}        & \textbf{3.83}  & \textbf{0.00}  & 3.6s  & \textbf{5.69}  & \textbf{0.00}  & 6.6s & 7.75  & 0.05  & 18.1s \\
SLIM (aug.) & 3.84  & 0.01  & 3.6s & 5.70  & 0.15  & 6.6s  & 7.84  & 1.17  & 18.1s \\
{ BOPO} (aug.) & \textbf{3.83} & \textbf{0.00} & 3.6s & 5.69 & 0.01 & 6.6s  & \textbf{7.75} & \textbf{0.04} & 18.1s\\ \midrule \bottomrule
\end{tabular}}
\vskip -0.1in
\end{table}

\begin{table}[tb!]
\centering
    \caption{Generalization on TSPLIB with various problem shapes.}
    \label{tb:tsp_gen}
\addtolength{\tabcolsep}{-3pt}
\vskip 0.15in
\resizebox{\linewidth}{!}{
\begin{tabular}{l||ccc|ccc|ccc|ccc}
\toprule
\midrule
\multirow{3}{*}{Method}  & \multicolumn{3}{c|}{$n<100$ } & \multicolumn{3}{c|}{$100\leq n<200$ } & \multicolumn{3}{c|}{$200\leq n<500$ }  & \multicolumn{3}{c}{$500\leq n<1k$} \\
& \multicolumn{3}{c|}{ (6 instances) } & \multicolumn{3}{c|}{ (21 instances) } & \multicolumn{3}{c|}{ (16 instances) }  & \multicolumn{3}{c}{ (6 instances) } \\
& Obj.$\downarrow$ & Gap$\downarrow$ & Time$\downarrow$  & Obj.$\downarrow$ & Gap$\downarrow$ & Time$\downarrow$  & Obj.$\downarrow$  & Gap$\downarrow$ & Time$\downarrow$   & Obj.$\downarrow$ & Gap$\downarrow$ & Time$\downarrow$   \\ \midrule
POMO (aug.)   & 6.26   & 2.36   & 0.15s   & 6.75   & 3.08   & 0.27s   & 10.63   & 14.81  & 0.95s  & 16.22  & 30.14  & 4.6s  \\
SLIM (aug.)   & 6.19   & 1.36   & 0.15s   & 6.88   & 5.24   & 0.27s   & 10.82   & 16.99  & 0.95s  & 19.40  & 55.57  & 4.6s  \\
{ BOPO} (aug.)   &\textbf{6.19}   & \textbf{1.26}   & 0.15s   &\textbf{6.72}   & \textbf{2.55}   & 0.27s   & \textbf{10.21}   & \textbf{10.41}  & 0.95s  & \textbf{15.29}  & \textbf{22.44}  & 4.6s  \\ \midrule \bottomrule
\end{tabular}}
\vskip -0.1in
\end{table}

\subsection{Flexible Job-shop Scheduling Problem}
\label{sec:res_fjsp} 

\textbf{Neural Model.}
For an FJSP instance represented as a disjunctive graph, we adopt the MGL model (see \cref{ap:nn}) to compute action probabilities, { which is similar to the model for JSP}.

\textbf{Training \& Test.} 
For evaluation, we use FJSP instances from the LA benchmark \cite{benchmark_la}. The benchmark includes \textit{e-data}, \textit{r-data}, and \textit{v-data}, where each operation can be allocated to 1-2 machines, 1-3 machines, and 1-$m$ machines, respectively. Following \citet{fjsp0}, we generate 25000 FJSP instances with 5 different shapes: \{$10\times5\times5$, $10\times10\times10$, $15\times10\times10$, $20\times5\times5$, $20\times10\times7$\}. For each shape, 2500 instances belong to \textit{r-data} and 2500 to \textit{v-data}. We set $B=128$, $K=16$, and $D=1$ during training, and employ both greedy rollout and sampling rollout with $B'$ solutions during inference.

\textbf{Baselines.}
For FJSP, we compare our proposal with both representative traditional PDRs and state-of-the-art RL-based constructive methods: DNN \cite{fjsp2} using the actor-critic framework, HG \cite{fjsp0} utilizing heterogeneous graphs for instance representation, and RS \cite{rs} employing residual scheduling to remove finished operations.

\textbf{Results on FJSP Benchmarks.} 
As shown in \cref{tb:fjsp}, with greedy rollout, { BOPO} significantly outperforms most baselines, only marginally falling behind RS on LA e-data. When all methods use sampling rollout with 100 solutions, the proposed method even achieves the best performance across all cases. Notably, increasing the number of sampled solutions consistently reduces the optimality gap.

\begin{table}[tb!]
\centering
    \caption{Average gaps (\%) on FJSP benchmarks.}\label{tb:fjsp}
\vskip 0.15in
\resizebox{\columnwidth}{!}{
\addtolength{\tabcolsep}{-4pt}
\begin{tabular}{c||ccc|ccc|c|ccc|cc}
\toprule
\midrule
  & \multicolumn{7}{c|}{Greedy Constructive}   & \multicolumn{5}{c}{Sampling Constructive}   \\ 
  & \multicolumn{3}{c}{Traditional PDR} & \multicolumn{3}{c}{RL} & { BOPO} & \multicolumn{3}{c}{$B'$=100} & $B'$=256 & $B'$=512 \\ \cmidrule(lr){2-4}\cmidrule(lr){5-7}\cmidrule(lr){8-8}\cmidrule(lr){9-11}\cmidrule(lr){12-12}\cmidrule(lr){13-13}
Benchmarks  & SPT    & MOR    & MWR    & DNN    & HG    & RS    & { BOPO} & HG    & RS    & { BOPO} & { BOPO} & { BOPO}  \\ \midrule \toprule
LA(e-data)  & 26.1   & 17.7   & 20.5   & 15.5   & 15.5  & 13.2  & 14.5 & 8.2   & 6.9   & 6.1 & 5.4 & \textbf{5.0} \\
LA(r-data)  & 28.7   & 14.4   & 17.8   & 12.1   & 11.2  & 9.6   & 8.4  & 5.8   & 4.7   & 4.0 & 3.6 & \textbf{3.4} \\
LA(v-data)  & 17.8   & 6.0    & 6.6    & 5.4    & 4.3   & 3.8   & 1.8  & 1.4   & 0.8   & 0.6 & 0.5 & \textbf{0.4} \\
\midrule \bottomrule

\end{tabular}}
\vskip -0.1in
\end{table}

\begin{table}[tb!]
\centering
\addtolength{\tabcolsep}{-4pt}
    \caption{Average gaps (\%) of various preference pair construction methods on the DMU benchmark.}
    \label{tb:jsp_pair}
\vskip 0.15in
\resizebox{\linewidth}{!}{
\begin{tabular}{c||c|ccc|c|c}
\toprule
\midrule
         & w/o Hybrid Rollout & \multicolumn{3}{c|}{w/o Uniform Filtering}    & w/o Best-anchored Pairing & \\ \cmidrule(lr){2-2} \cmidrule(lr){3-5} \cmidrule(lr){6-6}
   Shape & Sampling Rollout & Random  & Top-$K$ & Bottom-$K$  & Full Permutation Pairing  & { BOPO}  \\ \midrule \toprule
   20x15 & 11.6    & 12.0  & 12.8  & 12.5  & 11.9    & \textbf{10.9} \\
   20x20 & 13.0    & 12.9  & 14.5  & \textbf{12.5} & 13.3    & \textbf{12.5} \\
   30x15 & 15.2    & 14.0  & 15.9  & 14.6  & 15.5    & \textbf{13.3} \\
   30x20 & 16.7    & 16.4  & 18.5  & 16.7  & 16.6    & \textbf{15.8} \\ 
   40x15 & 12.1    & 11.7  & 13.2  & 11.8  & \textbf{11.2}   & \textbf{11.2} \\
   40x20 & 17.5    & \textbf{16.1} & 18.7  & 16.4  & 16.5    & \textbf{16.1} \\
   50x15 & 12.1    & 12.1  & 12.8  & 11.4  & 11.2    & \textbf{10.9} \\
   50x20 & 17.1    & \textbf{16.3} & 18.8  & 16.6  & 16.4    & \textbf{16.3} \\ \midrule
   Avg   & 14.4    & 14.0  & 15.6  & 14.1  & 14.1    & \textbf{13.3} \\ \midrule \bottomrule
\end{tabular}}
\vskip -0.1in
\end{table}

\subsection{Ablation Study}
\label{sec:res_ab}

\begin{figure*}[ht!]
\vskip 0.2in
    \centering
    \begin{subfigure}[b]{0.49\textwidth} 
        \includegraphics[width=\textwidth]{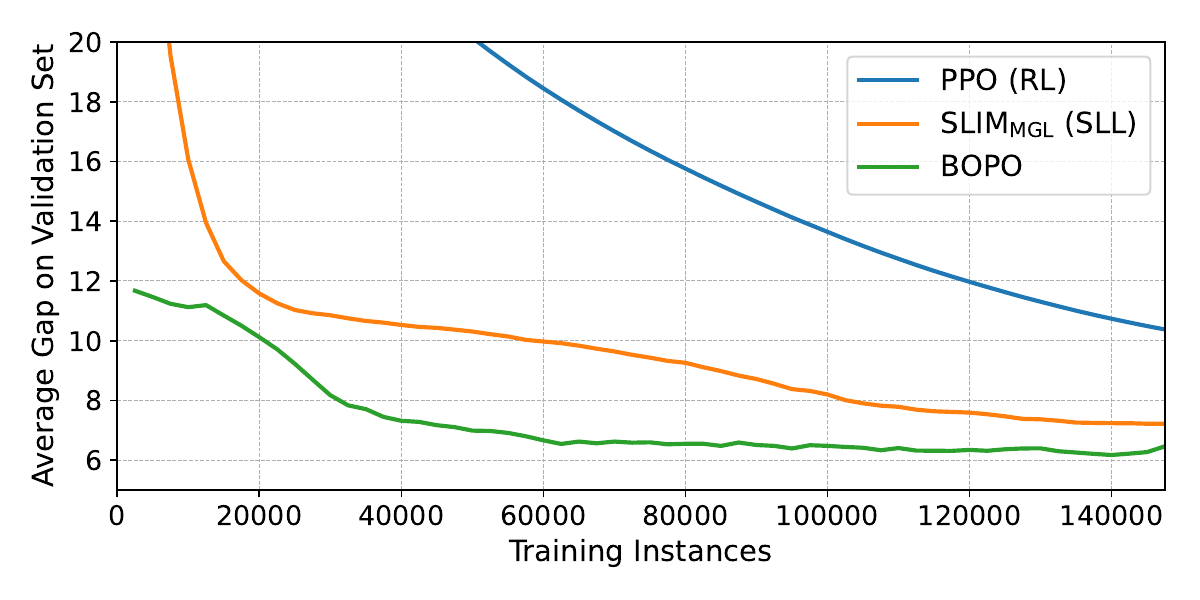}
        \caption{}
        \label{fg:jsp_curve}
    \end{subfigure}
    \begin{subfigure}[b]{0.49\textwidth}
        \includegraphics[width=\textwidth]{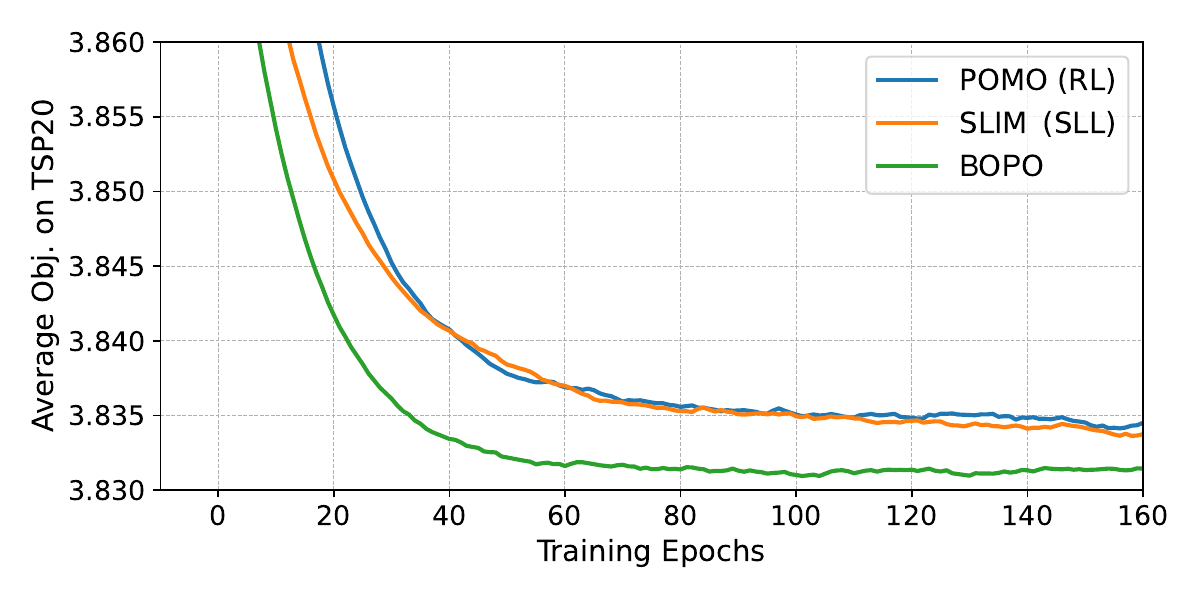} 
        \caption{}
        \label{fg:tsp_curve}
    \end{subfigure}
    \caption{Training curves of different training paradigms for: (a) JSP and (b) TSP.}
    \label{fg:train_curve}
\vskip -0.2in
\end{figure*}

\textbf{Higher Sample Efficiency of the { BOPO} Training Paradigm.} We compare { BOPO} with two representative training paradigms: RL and SLL. For JSP, we compare { BOPO} with RL-based PPO and SLL-based SLIM$_\text{MGL}$, all using the MGL model and identical training settings. For TSP, we compare { BOPO} with RL-based POMO and SLL-based SLIM, all using the POMO model and identical training settings. The training curves in \cref{fg:train_curve} demonstrate our proposal's higher sample efficiency, achieving lower optimality gaps than both baselines with the same number of training instances, with the advantage being more pronounced when training data is limited.

\textbf{Effectiveness of the { Best-anchored} Preference Pair Construction Method.} To validate our three-step preference pair construction method, we replace each step with a simpler alternative. We substitute sampling rollout for hybrid rollout, replace uniform filtering with random, top-$K$, or bottom-$K$ filtering, and use full permutation pairing instead of best-anchored pairing while maintaining the same total number of pairs. As shown in \cref{tb:jsp_pair}, { BOPO}'s performance deteriorates without (w/o) any of these components, demonstrating the necessity of our design. Notably, top-$K$ filtering leads to significantly worse performance, highlighting the importance of filtering worse solutions to create sufficient preference differences in pair construction. Additional analyses of hybrid rollout effectiveness are provided in \cref{ap:hybrid}. 

\textbf{Superiority of the { Objective-guided Preference Optimization} Loss.} To evaluate our proposed loss with a preference-based scaling factor, we compare it with other preference optimization losses: the classic DPO loss ($\mathcal L_{DPO}$) \cite{dpo}, the popular SimPO loss ($\mathcal L_{SimPO}$) \cite{simdpo}, and a variant of { BOPO} loss without the scaling factor ($\mathcal L_{BOPO^-}$). Complete loss formulations are provided in \cref{ap:other_methods}. For DPO, which requires a reference model to prevent excessive policy deviation, we use the old model from 10 episodes prior, similar to PPO, with the common hyperparameter setting $\beta=0.1$. For SimPO, we follow their standard parameters with $\beta=2$ and $\gamma=1$. As shown in \cref{tb:jsp_loss}, { BOPO} achieves the best performance across all benchmarks for both $B'=128$ and $B'=512$, with particularly significant improvements on the DMU benchmark, demonstrating the effectiveness of our loss design and its preference-based scaling factor. {  It is worth noting that SimPO performs poorly on DMU, where the test distribution diverges from training. This hints at potential overfitting caused by its target margin term $\gamma$.}

\begin{table}[tb!]
\centering
    \caption{Average gaps (\%) of various loss functions for JSP.}
    \label{tb:jsp_loss}
\vskip 0.15in
\resizebox{\linewidth}{!}{
\addtolength{\tabcolsep}{-4pt}
\begin{tabular}{c||cccc|cccc}
\toprule
\midrule
          & \multicolumn{4}{c|}{$B'=128$} & \multicolumn{4}{c}{$B'=512$}  \\ \cmidrule(lr){2-5} \cmidrule(lr){6-9}
Benchmark & $\mathcal L_{SimPO}$ & $\mathcal L_{DPO}$ & $\mathcal L_{BOPO^-}$  & $\mathcal L_{BOPO}$ & $\mathcal L_{SimPO}$ & $\mathcal L_{DPO}$ & $\mathcal L_{BOPO^-}$  & $\mathcal L_{BOPO}$  \\ \midrule \toprule
TA        & 8.5 & 8.7 & 8.5 & 8.2 & 7.7  & 7.8  & 7.6 & \textbf{7.5}  \\
LA        & 2.9 & 2.9 & 2.9 & 2.9 & 2.4  & 2.5  & 2.4 & \textbf{2.3}  \\
DMU       & 15.2& 14.5& 14.1& 13.8& 14.1 & 13.6 & 13.2& \textbf{12.9} \\
\midrule \bottomrule
\end{tabular}}
\vskip -0.1in
\end{table}

\begin{figure*}[ht!]
\vskip 0.2in
    \centering
    \begin{subfigure}[b]{0.3\textwidth} 
        \includegraphics[width=\textwidth]{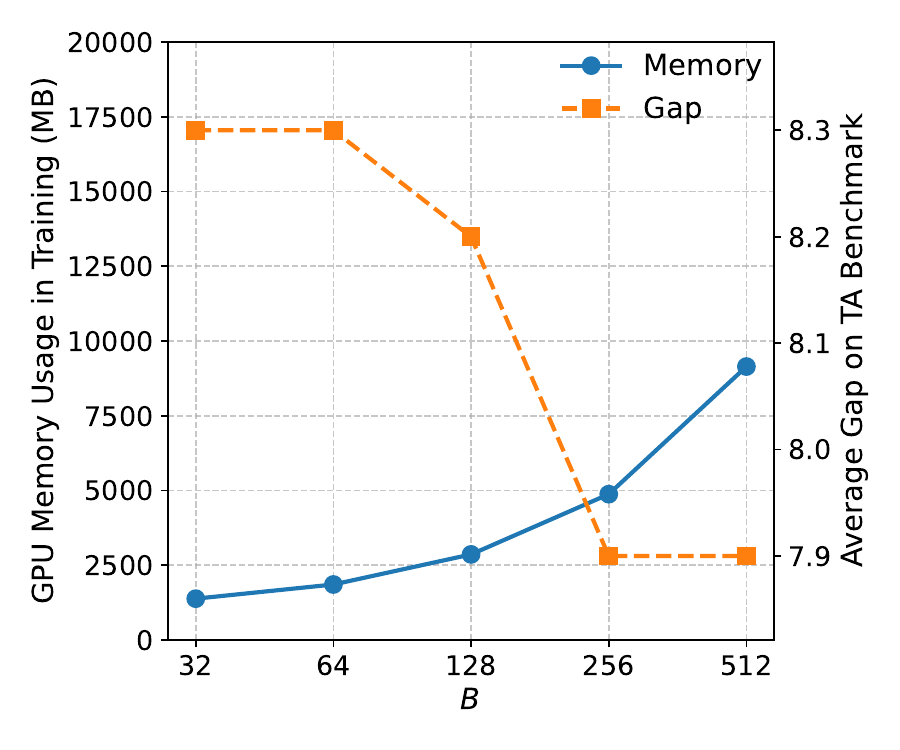}
        \caption{}
        \label{fg:b_para}
    \end{subfigure}
    \hfill
    \begin{subfigure}[b]{0.3\textwidth}
        \includegraphics[width=\textwidth]{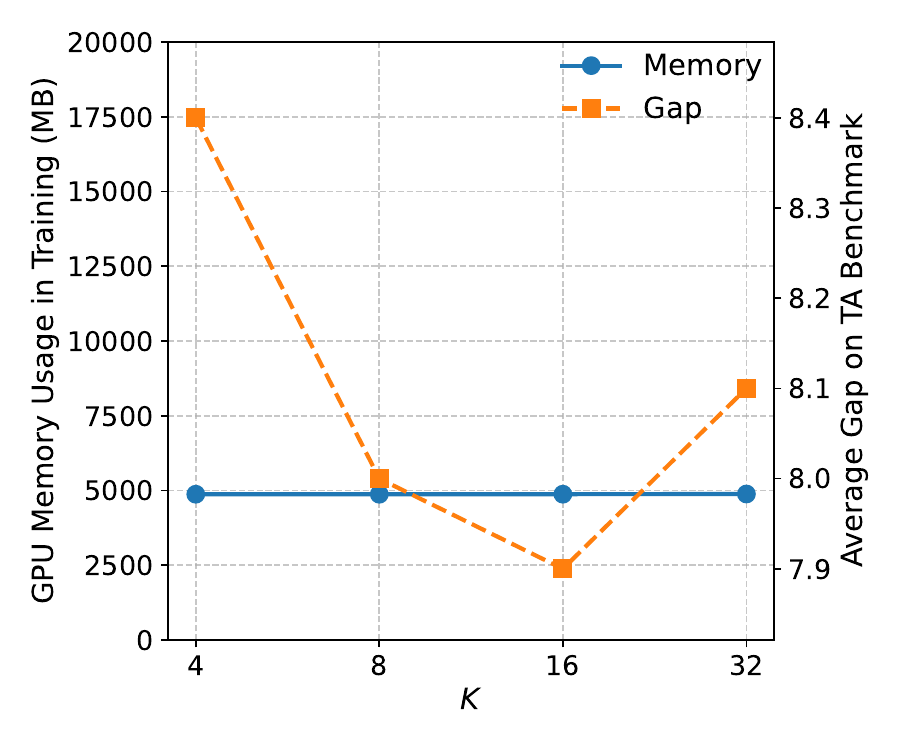} 
        \caption{}
        \label{fg:k_para}
    \end{subfigure}
    \hfill
    \begin{subfigure}[b]{0.3\textwidth}
        \includegraphics[width=\textwidth]{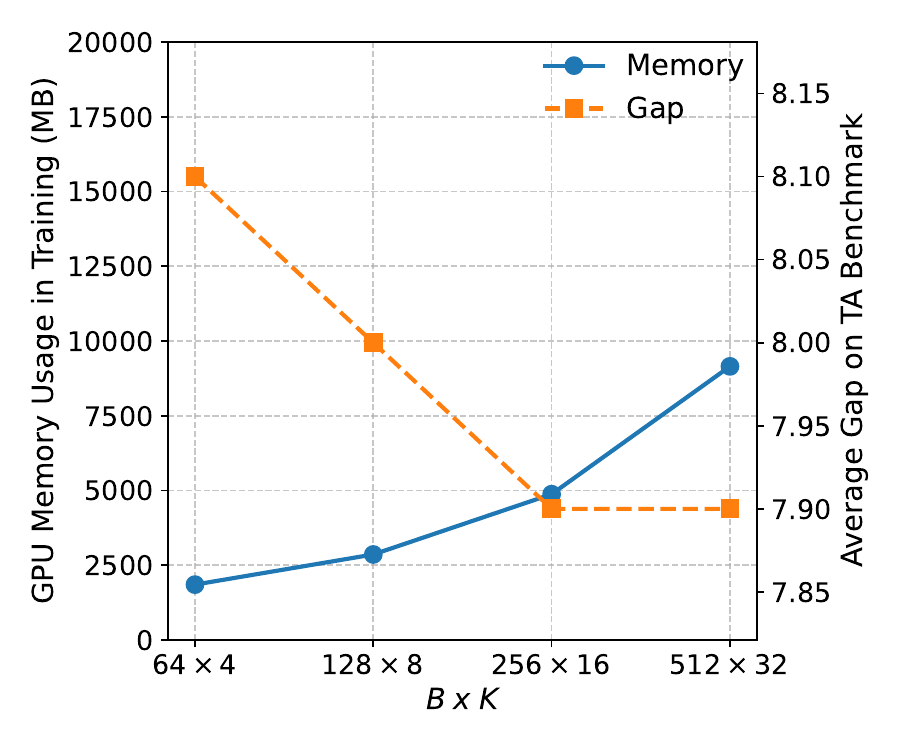}
        \caption{}
        \label{fg:bk_para}
    \end{subfigure}
    \caption{GPU memory usage in training and average gap (\%) on TA benchmark for: (a) varying $B$, (b) varying $K$, and (c) varying $B\!\times\!K$.}
    \label{fg:para}
\vskip -0.2in
\end{figure*}

\subsection{Hyperparameter Study}

{  BOPO} has two crucial hyperparameters: the solution number of hybrid rollouts $B$ and the number of filtered solutions $K$. We analyze their individual effects by varying $B\in\{32,64,128,256,512\}$ and $K\in\{4,8,16,32\}$. Additionally, we explore their interaction by maintaining a fixed ratio $B/K=16$ while scaling both parameters $K\times B\in\{4\times64,8\times128,16\times256,32\times512\}$.

\textbf{Effect of the Solution Number of Hybrid Rollouts.} As shown in \cref{fg:b_para}, increasing the number of sampled solutions during training, i.e., larger $B$, improves solution quality by enhancing the probability of collecting higher-quality solutions. However, GPU memory consumption grows with $B$ due to parallel computation, raising costs. While the performance improves significantly up to $B=256$, further increases yield diminishing returns despite rising computational costs, making $B=256$ a reasonable choice. {  We also evaluate our method on TSP20/50 with $B=20/50$, the performance remains comparable to $B=128$ (detailed in \cref{ap:tsp}). This suggests that for small-scale problems, the optimal $B$ is low since the model can efficiently sample high-quality solutions.}

\textbf{Effect of the Number of Filtered Solutions.} The number of filtered solutions $K$ determines the number of preference pairs, making it a critical parameter. As shown in \cref{fg:k_para}, while increasing $K$ generates more preference pairs, it also increases the similarity among solutions, as they are more likely to come from the same local region. Our experiments show that $K\!=\!16$ achieves the best performance, with either larger or smaller values leading to performance degradation. This suggests that a moderate $K$ value balances the trade-off between sufficient training data and solution diversity.

\textbf{Effect of Interaction Between Number of Rollouts and Filtered Solutions.} $B$ and $K$ are interdependent parameters, as $B$ affects the uniform filtering step size $\lfloor B/K \rfloor$, which influences the similarity between solutions. As shown in \cref{fg:bk_para}, we maintain a fixed step size of 16 while scaling $B$ and $K$ proportionally. The performance improves until $B\times K$ reaches $256\times 16$, after which larger values yield minimal gains despite increased memory costs. This further validates our choice of $B=256$ and $K=16$ as recommended parameters.

\section{Conclusion}
\label{sec:conclusion} 
In this work, we present { BOPO}, a preference optimization-based training paradigm for neural combinatorial optimization. By introducing { best-anchored} preference pair construction and a novel { objective-guided pairwise} loss function for COPs, our proposal achieves higher sample efficiency than mainstream RL and recent SLL paradigms. Extensive experiments on JSP, TSP, and FJSP demonstrate { BOPO}'s superior performance over state-of-the-art neural constructive methods, while requiring significantly less time to deliver solutions competitive with traditional problem-specific iterative heuristics. The proposed method requires neither expensive labels nor specialized design of the Markov Decision Process, making it easy to use in practice. More importantly, it establishes a general training paradigm that can be readily applied to various neural models for solving different COPs.

Although { BOPO} demonstrates superior sample efficiency compared with RL and SLL, one limitation is that it still requires a relatively large number of rollout solutions, similar to SLIM. This incurs moderate costs in collecting high-quality solutions for effective model learning. In future work, we will explore efficient ways to obtain high-quality solutions that facilitate model training. One promising direction is to leverage problem invariance and solution symmetry to efficiently generate diverse high-quality training data. Another direction is to efficiently enhance solution quality by incorporating problem-specific heuristics during training, providing better learning signals for the model.

\section*{Acknowledgements}
{ This work is supported by the National Natural Science Foundation of China (62472461) and the Guangdong Basic and Applied Basic Research Foundation (2024A1515010871, 2025A1515010129).}

\section*{Impact Statement}

This paper introduces research aimed at advancing machine learning, poised to facilitate industries and enhance decision-making. However, its adoption must be accompanied by careful consideration of ethical, societal, and environmental implications to ensure responsible and equitable use.

\bibliography{refs}
\bibliographystyle{icml2025}

\newpage
\appendix

\onecolumn

\section{Formalization of Problems}
\label{ap:problem}
\subsection{Job-shop Scheduling Problems}
JSP consists of a set of jobs $\mathcal J=\{J_1, \cdots, J_n\}$, a set of operations $\mathcal O=\{O_1, \cdots, O_l\}$, and a set of machines $\mathcal M=\{M_1, \cdots, M_m\}$. Each job $J_i \in \mathcal J$ is composed of a sequence of $m$ operations $(O_{l_{i1}}, \cdots, O_{l_{im}})$, where $l_{ij} \in \{1, \cdots, l\}$. It must be completed sequentially in a strict order. Each operation $O_j \in \mathcal O$ must be performed on a specific machine $M_j \in \mathcal M$ continuously for $t_j$ seconds, and each machine can only process one operation at a time. For convenience, we define assignable operations at step $t$ as $\mathcal O_t$, the pending operation of job $J_i$ at step $t$ as $O_{(i,t)}$, and the corresponding machine of $O_{(i,t)}$ as $M_{(i,t)}$. Once a scheduling plan is determined, the completion time $C(O_j)$ of each operation $O_j$ is decided accordingly, resulting in the maximum completion time $C(\bm y)=\max_{i\in\{1,\dots,n\}} C(J_i)$ of all jobs (i.e., makespan), where $C(J_i)=\max_{j\in\{l_{i1},\dots,l_{im}\}} C(O_j)$ represents the completion time of job $J_i$. Makespan is typically the objective to be minimized in JSP.

\textbf{Disjunctive Graph}. JSP can be represented using a disjunctive graph $G=(V,A,E)$. In this graph, the node set $V=\{ O_j\; |\; O_j\in\mathcal O \}$ represents operations, the directed edge set $A=\{ O_{l_{ij}}\rightarrow O_{l_{i(j+1)}} \;|\; O_{l_{ij}},O_{l_{i(j+1)}}\in\mathcal O \}$ indicates precedence constraints between successive operations ($O_{l_{ij}} \rightarrow O_{l_{i(j+1)}}$), and the disjunctive (undirected) edge set $E$ connects operations performed on the same machine. A feasible solution is obtained by assigning directions to the undirected edges in $E=\{ O_{j} \leftrightarrow O_{j^\prime} \;|\; O_j,O_{j^\prime}\in\mathcal O, M_{j} = M_{j^\prime} \}$, resulting in a directed acyclic graph. 

\textbf{Features of JSP.} Following previous works \cite{l2d,slim}, we define two types of features. The first type consists of static state features $\bm{s}_j$ assigned to each operation $O_j$ in the node set $V$, while the second type comprises contextual features $\bm{c}_i$ based on job $J_i$ information at step $t$. The state features $\bm{s}_j$ are fed into the encoder to compute embeddings for each node (operation), while the contextual features $\bm{c}_i$ are combined with these embeddings to jointly contribute to the prediction.
\cref{tb:fea_jsp_s} details the state features $\bm{s}_j \in \mathbb{R}^{15}$, and \cref{tb:fea_jsp_c} details the contextual features $\bm c_i\in \mathbb{R}^{11}$.

\begin{table}[bt!]
    \centering
\small
\caption{The state features $\bm{s}_j \in \mathbb{R}^{15}$ that describes the information of an operation $O_j$ for JSP. }
\label{tb:fea_jsp_s}
\vskip 0.15in
\begin{tabular}{r|l}
\midrule
ID  & Description \\ \midrule
1 & The processing time $t_j$ of the operation. \\
2 & The completion of job $J_i$ up to $O_j$: $\sum_{j'=l_{i1}}^{j} t_{j'} / \sum_{j'\in \mathcal{O}_i} t_{j'}$.   \\
3 & The completion of job $J_i$ after $O_j$: $\sum_{j'=j+1}^{l_{im}} t_{j'} / \sum_{j'\in \mathcal{O}_i} t_{j'}$. \\
4-6   & The $1^{st}$, $2^{nd}$, and $3^{rd}$ quartile among processing times of operations on job $J_i$.  \\
7-9   & The $1^{st}$, $2^{nd}$, and $3^{rd}$ quartile among processing times of operations on machine $M_j$.  \\
10-12 & The difference between $t_j$ and feature 4~6.   \\
13-15 & The difference between $t_j$ and feature 7~9. \\ \midrule
\end{tabular}
\end{table}

\begin{table}[bt!]
\centering
\small
\caption{The context features $\bm{c}_i\in \mathbb{R}^{11}$ that describes the status of a job $J_i$ within a partial solution $\bm{y}_{<t}$ at step $t$ for JSP. }
\label{tb:fea_jsp_c}
\vskip 0.15in
\begin{tabular}{r|l}
\midrule
ID  & Description \\ \midrule
1   & $C(J_i)$ minus the completion time of machine $M_{(i,t)}$  \\ 
2   & $C(J_i)$ divided by the makespan of partial solution $C(\bm{y}_{<t})$. \\ 
3   & $C(J_i)$ minus the average completion time of all jobs.  \\ 
\multirow{2}{*}{4-6}   & The difference between $C(J_i)$ and the $1^{st}$, $2^{nd}$, and $3^{rd}$ quartile computed among the completion  \\ 
    & time of all jobs.    \\ 
7   & The completion time of machine $M_{(i,t)}$ divided by the makespan of the partial solution $C(\bm{y}_{<t}).$ \\ 
8   & The completion time of machine $M_{(i,t)}$ minus the average completion of all machines.  \\ 
\multirow{2}{*}{9-11}   & The difference between the completion of $M_{(i,t)}$ and the $1^{st}$, $2^{nd}$, and $3^{rd}$ quartile computed  \\
    & among the completion time of all machines.  \\ \midrule
\end{tabular}
\end{table}

\subsection{Traveling Salesman Problem}
Two-dimensional Euclidean TSP, which is discussed in this paper, involves $n$ nodes, where each node $i\in\{1,\dots,n\}$ is represented by a two-dimensional coordinate, forming a fully connected graph. The distance $C(i,j)$ between nodes $i$ and $j$ is calculated by their coordinates. The objective of TSP is to find the shortest Hamiltonian cycle $g(\bm y)=\sum_{j=1}^nC(y_j,y_{j+1})$, where $\bm y=(y_1,\dots,y_n,y_1),y_j \in \{1,\dots,n\}$ visits each node exactly once and returns to the starting node. 

\textbf{Features of TSP.} Similar to \citet{tsp_am}, the input of TSP consists of $n$ nodes with 2-dimensional features. At each decoding step, the context embedding $\bm{h}_c$ is defined as the concatenation of embeddings from the first and last visited nodes. The coordinates of each instance are sampled from a uniform distribution on $[0,1]^2$.

\subsection{Flexible Job-shop Scheduling Problem}
FJSP, a variant of JSP, better reflects real-world scenarios. Unlike JSP, FJSP allows each operation $O_{j}\in\mathcal O$ to choose from multiple candidate machines $\mathcal M_j\subseteq \mathcal M$, rather than being restricted to a specific one. We redefine the processing time of $O_{j}$ in machine $M_k\in\mathcal M_j$ as $t_{jk}$, and denote the $t_j=\frac{1}{|\mathcal M_j|}\sum_{k=1}^{|\mathcal M_j|}t_{jk}$ as the average processing time of operation $O_j$.
This flexibility significantly increases decision-making complexity, leading to a denser disjunctive graph.

\textbf{Disjunctive Graph for FJSP}. To simplify the disjunctive graph for FJSP, we introduce operation-machine nodes, where an operation is decomposed into multiple operation-machine nodes. Each operation-machine node is similar to a node in JSP and can be treated as an action. Specifically, the disjunctive graph is formalized as $G=(V,A,E,U)$, where $V=\{ (O_j, M_k)\; |\; O_j\in\mathcal O, M_k\in\mathcal M_j \}$ represents the set of all operation-machine pairs, $A=\{ (O_{l_{ij}}, M_k)\rightarrow(O_{l_{i(j+1)}}, M_{k^\prime}) \;|\; O_{l_{ij}},O_{l_{i(j+1)}}\in\mathcal O; M_k\in\mathcal M_{l_{ij}}; M_{k^\prime}\in\mathcal M_{l_{i(j+1)}} \}$ denotes the directed edge set, $E=\{ (O_{j}, M_k)\leftrightarrow(O_{j^\prime}, M_{k}) \;|\; O_j,O_{j^\prime}\in\mathcal O; M_k\in\mathcal M_{j}\cap\mathcal M_{j^\prime} \}$ denotes the disjunctive (undirected) edge set, and $U=\{ (O_{j}, M_k)\leftrightarrow(O_{j}, M_{k^\prime}) \;|\; O_j\in\mathcal O,; M_k, M_{k^\prime}\in\mathcal M_{j} \}$ connects all operation-machine nodes belonging to the same operation.

\textbf{Features of FJSP.} Similar to JSP, we define two types of features for FJSP. To accommodate operation-machine nodes used in FJSP, we split the context features into job context features $\bm c^J_i$ and machine context features $\bm c^M_k$, where $\bm c^J_i$ represents the context features of job $J_i$, and $\bm c^M_k$ represents the context features of machine $M_k$. Similarly, the static state features $\bm s_j\rightarrow \bm s_{jk}$ are modified from describing operation $O_j$ to describing the operation-machine nodes $(O_j, M_k)$. 
\cref{tb:fea_fjsp_s} details the state features $\bm{s}_{jk} \in \mathbb{R}^{15}$, and \cref{tb:fea_fjsp_c} details the contextual features $\bm c^J_i,\bm c_k^M\in \mathbb{R}^{5}$.

\begin{table}[tb!]
    \centering
\small
\caption{The state features $\bm{s}_{jk} \in \mathbb{R}^{15}$ that describes the information of an operation-machine node $(O_j, M_k)$ for FJSP.}
\label{tb:fea_fjsp_s}
\vskip 0.15in
\begin{tabular}{r|l}
\midrule
ID  & Description \\ \midrule
1 & The processing time $t_{jk}$ of the node $(O_j,M_k)$. \\
2 & The average completion of job $J_i$ up to $O_j$: $\sum_{j'=l_{i1}}^{j} t_{j'} / \sum_{j'\in \mathcal{O}_i} t_{j'}$.   \\
3 & The average completion of job $J_i$ after $O_j$: $\sum_{j'=j+1}^{l_{im}} t_{j'} / \sum_{j'\in \mathcal{O}_i} t_{j'}$. \\
4-6   & The $1^{st}$, $2^{nd}$, and $3^{rd}$ quartile among processing times of operations on job $J_i$.  \\
7-9   & The $1^{st}$, $2^{nd}$, and $3^{rd}$ quartile among processing times of operations on machine $M_k$.  \\
10-12 & The difference between $t_j$ and feature 4-6.   \\
13-15 & The difference between $t_j$ and feature 7-9. \\ \midrule
\end{tabular}
\end{table}

\begin{table}[tb!]
\centering
\small
\caption{The job context features $\bm c^J_i\in \mathbb{R}^{5}$ and the machine context features $\bm c^M_k\in \mathbb{R}^{5}$ that describe the status of a job $J_i$ and a machine $M_k$ within a partial solution $\bm{y}_{<t}$ at step $t$ for JSP. }
\label{tb:fea_fjsp_c}
\vskip 0.15in
\begin{tabular}{r|l}
\toprule \midrule
Job ID  & Description \\ \midrule
1   & $C(J_i)$ divided by the makespan of partial solution $C(\bm{y}_{<t})$. \\ 
2   & $C(J_i)$ minus the average completion time of all jobs.  \\ 
\multirow{2}{*}{3-5}   & The difference between $C(J_i)$ and the $1^{st}$, $2^{nd}$, and $3^{rd}$ quartile computed among the completion  \\   
    & time of all jobs.    \\  \midrule \toprule
Machine ID  & Description \\ \midrule
1   & The completion time of machine $M_k$ divided by the makespan of the partial solution $C(\bm{y}_{<t}).$ \\ 
2   & The completion time of machine $M_k$ minus the average completion of all machines.  \\ 
\multirow{2}{*}{3-5}   & The difference between the completion of $M_k$ and the $1^{st}$, $2^{nd}$, and $3^{rd}$ quartile computed  \\
    & among the completion time of all machines.  \\ \midrule
\end{tabular}
\end{table}

\section{Neural Model for Scheduling Problems}
\label{ap:nn}

For scheduling problems, we design an efficient neural model named MGL, which combines a multi-layer graph attention network (GAT) \cite{gat} encoder with a long short-term memory (LSTM) based \cite{lstm} context attention decoder.

\subsection{Neural Model for JSP}

\textbf{Encoder.}
Since the disjunctive graph contains two types of edges: directed edges related to jobs, and disjunctive edges (undirected) related to machines, we treat it as a two-layer graph to better distinguish between these two edge types, i.e., $G_{job}=(V,A),G_{mac}=(V,E)$, and $G=G_{job}\cup G_{mac}$. One layer contains only directed edges, while the other contains only disjunctive edges. To process this structure, we introduce a multi-layer GAT as the encoder, where each layer can be considered a standard 2-head GAT. The computation for an $N$-layer $\text{GAT}^N$ is as follows:
$$\text{GAT}^N(\bm x,G_1,\dots,G_N)= [ \sigma(\text{GAT}_1(\bm x,G_1))||\cdots||\sigma(\text{GAT}_N(\bm x,G_N))].$$
In our encoder, we stack two 2-layer GAT as follows to embed $\bm e_j$ of operation $O_j$:
$$\bm e_j=[\bm s_j || \sigma(\text{GAT}_{second}^2([\bm s_j || \sigma(\text{GAT}_{first}^2(\bm s_j, G_{job}, G_{mac}))], G_{job}, G_{mac}))].$$

\textbf{Decoder.}
At step $t$, we use the embedding $\bm e_{y_{t-1}}$ of the operation selected at step $t-1$ as input to the LSTM, which computes the query $\bm q_t$ as follows:
$$\bm q_t=\text{LN}(\text{LSTM}(\sigma(\bm e_{y_{t-1}}\cdot W_1)))\cdot W_2.$$
For each operation $O_j$, we concatenate context feature $\bm{c}_i$ of its job $J_i$ with the embedding $\bm e_j$ to obtain the key $\bm k_{t,j}$ as follows:
$$\bm k_{t,j}=[\sigma(\bm{c}_i \cdot W_3) ||\bm  e_{j}] \cdot W_4.$$
Finally, the query $\bm q_t$ attends to the keys $\bm k_{t,j}$ of assignable operations, computing the attention to generate the policy distribution for action selection:
$$\pi(O_j|t)=\frac{\exp(\bm q_t\cdot \bm k_{t,j}^\top)}{\sum_{j'\in \mathcal O_t} \exp(\bm q_t\cdot \bm k_{t,j'}^\top) }.$$

\subsection{Neural Model for FJSP}
\textbf{Encoder for FJSP.}
We also employ the MGL model for FJSP. With the introduction of operation-machine nodes and the operation-related edge set $U$, we redefine $G_{opr}=(V,U)$  and $G=G_{job} \cup G_{opr} \cup G_{mac}$.
 The following describes the embedding $\bm e_{jk}$ of a 3-layer GAT to node $(O_j,M_k)$:
 $$\bm e_{jk}=[\bm s_{jk} || \sigma(\text{GAT}_{second}^3([\bm s_{jk} || \sigma(\text{GAT}_{first}^3(\bm s_{jk}, G_{job}, G_{mac}, G_{opr}))], G_{job}, G_{mac}, G_{opr}))].$$

\textbf{Decoder for FJSP.}
For operation-machine node $(O_j,M_k)$ of job $J_i$, the key $\bm{k}_{t,j,k}$ is modified as:
$$\bm k_{t,j,k}=[\sigma(\bm{c}^J_i \cdot W_3)|| \sigma(\bm{c}^M_k \cdot W_4) ||\bm  e_{jk}] \cdot W_5.$$
Due to the fact that query $\bm q_t$ is independent of operation-machine nodes, the policy distribution for nodes $(O_j,M_k)$ is:
$$\pi((O_j,M_k)|t)=\frac{\exp(\bm q_t\cdot \bm k_{t,j,k}^\top)}{\sum_{O_{j'}\in \mathcal O_t,M_{k'}\in\mathcal M_{j'}} \exp(\bm q_t\cdot \bm k_{t,j',k'}^\top) }.$$

\subsection{Comparison with Neural Model Used in SLIM}
Our MGL is compared with the neural model used in SLIM \cite{slim}, denoted as GAT-MHA.
The main distinction between these models lies in their decoders, significantly impacting memory consumption and computational efficiency, as analyzed in \cref{tb:nn_ana}. All data are collected from a $15\times15$ instance, with the solution size $B$ set to 256.

The multi-head attention (MHA) module in GAT-MHA introduces additional weight matrices that contribute to its higher parameter count. In contrast, LSTM in MGL achieves efficiency through recurrent weight sharing. GAT-MHA requires 32.8\% more memory during forward propagation than MGL. Moreover, GAT-MHA's backward pass memory usage is \textbf{27× higher} than MGL's, primarily due to MHA's gradient computation needs: \textit{Intermediate Activation Storage} and \textit{Gradient Scaling with Heads}. For the former, MHA must retain attention score matrices and head-specific outputs during forward pass for gradient computation, while LSTM's recurrent nature minimizes intermediate storage. For the latter, MHA's memory overhead scales linearly with the number of attention heads, as gradients for each head's parameters are stored separately.  

\begin{table}[tb!]
\centering
\small
\caption{MGL's efficiency compared with SLIM's model. }
\label{tb:nn_ana}
\vskip 0.15in
\begin{tabular}{l|l|l|l|l}
\toprule \midrule
\multirow{2}{*}{Model} & \multirow{2}{*}{Parameters} & Memory Usage (MB)   & Training Time (ms) & Inference Time (ms) \\
                       &                             & During Training     & per Instance       & per Instance        \\ \midrule
MGL                    & 351.23K                     & 262.19              & 1395.99            & 800.04              \\
GAT-MHA                & 376.96K (+25.73K)           & 7110.69 (27x)       & 1785.7 (+27.9\%)   & 851.81 (+6.4\%)     \\
\bottomrule
\end{tabular}
\vskip -0.1in
\end{table}

\section{Comparison with Other Loss Functions}
\label{ap:other_methods}

\subsection{Formulations of Loss Functions}

In SLIM \cite{slim}, the locally optimal solution $y_o$ is treated as a pseudo-label, and the model is trained using cross-entropy loss. The loss function of SLIM can be expressed as::
\begin{equation}
\mathcal L_{SLIM}(\pi_{\bm\theta}|\bm x,\bm{y}_o)=-\frac{1}{|\bm{y}_o|}\log\pi_{\bm\theta}(\bm{y}_o|\bm x).
\end{equation}
This formulation effectively maximizes the average log-likelihood of the locally optimal solution. In contrast to SLIM, our method incorporates suboptimal solutions into the loss function, effectively minimizing the average log-likelihood of these suboptimal candidates.
This approach improves sample efficiency and accelerates the convergence of model training.

DPO \cite{dpo} employs a reference model $\pi_\text{ref}$, analogous to RLHF \cite{rlhf}, to regularize the trained model against excessive deviation from the initial policy. The DPO loss function is defined as:  
\begin{equation}  
\mathcal{L}_{\text{DPO}}(\pi_{\bm\theta} | \pi_\text{ref}, \bm x, \bm{y}_w, \bm{y}_l) =  
-\log \sigma \bigg(  
\beta \log \frac{\pi_{\bm\theta}(\bm{y}_w | \bm x)}{\pi_\text{ref}(\bm{y}_w | \bm x)}  
- \beta \log \frac{\pi_{\bm\theta}(\bm{y}_l | \bm x)}{\pi_\text{ref}(\bm{y}_l | \bm x)}  
\bigg),  
\end{equation}  
where $\beta$ controls the strength of regularization toward the reference model $\pi_\text{ref}$. In contrast, our method eliminates the dependency on an explicit reference model, simplifying the training framework while avoiding potential distributional shift issues.  

To simplify the training phase and align the training goal with the generation goal, SimPO \cite{simdpo} eliminates the reference model and simplifies the loss function as follows:
\begin{equation}\begin{aligned}
\mathcal L_{SimPO}(\pi_{\bm\theta}|\bm x,\bm{y}_w,\bm{y}_l) = -\log\sigma 
\bigg(
  \frac{\beta}{|\bm{y}_w|}\log \pi_{\bm\theta}(\bm{y}_w|\bm x)
- \frac{\beta}{|\bm{y}_l|}\log \pi_{\bm\theta}(\bm{y}_l|\bm x)
- \gamma
\bigg),
\end{aligned}\end{equation}
where $\beta$ is a constant that controls the scaling of the difference, and $\gamma$ is a target margin term. In contrast, {  BOPO} uses an adaptive objective gap factor to scale the differences, instead of relying on additional hyperparameters.

{ 
Considering the similarity between the BOPO loss and that of policy gradients, we additionally analyze the loss function in the REINFORCE algorithm \cite{tsp_pomo} here:
\begin{equation}\begin{aligned}
\mathcal L_{PG}(\pi_{\bm\theta}|\bm x,\bm{y}) = 
-(g(\bm y) - b)\log \pi_{\bm\theta}(\bm{y}_w|\bm x),
\end{aligned}\end{equation}
where $b$ is a baseline to distinguish positive or negative optimization signals for each sample $\bm y$. In POMO, $b= \sum_i^B g(\bm y_i) / B$ is the average objective of $B$ samples. Note that, both BOPO and REINFORCE use the exact reward (objective) to optimal policy. However, BOPO directly distinguishes optimization signals based on exact preferences, and the strength of the optimization signal correlates with the difference in log-likelihood, requiring the model to maximize the probabilities gap between $\bm y_w$ and $\bm y_l$. 
}

\subsection{Gradient Analysis}
Let \(z\) denote the argument of the sigmoid in \cref{loss}:
\[
z = \frac{g(\bm{y}_l)}{g(\bm{y}_w)} \left( \frac{1}{|\bm{y}_w|} \log \pi_\theta(\bm{y}_w|\bm{x}) - \frac{1}{|\bm{y}_l|} \log \pi_\theta(\bm{y}_l|\bm{x}) \right).
\]
The gradient of \(\mathcal{L}_{\text{BOPO}}\) with respect to \(\theta\) is:
\[
\nabla_\theta \mathcal{L}_{\text{BOPO}} = \frac{\partial \mathcal{L}_{\text{BOPO}}}{\partial z} \cdot \nabla_\theta z.
\]

Derivative of \(-\log \sigma(z)\) becomes:
   \[
   \frac{\partial \mathcal{L}_{\text{BOPO}}}{\partial z} = -(1 - \sigma(z)).
   \]
   
Gradient of \(z\) with respect to \(\theta\):
   \[
   \nabla_\theta z = \frac{g(\bm{y}_l)}{g(\bm{y}_w)} \left( \frac{1}{|\bm{y}_w|} \nabla_\theta \log \pi_\theta(\bm{y}_w|\bm{x}) - \frac{1}{|\bm{y}_l|} \nabla_\theta \log \pi_\theta(\bm{y}_l|\bm{x}) \right).
   \]

Combining these, the total gradient becomes:
\[
\nabla_\theta \mathcal{L}_{\text{BOPO}} =  \underbrace{-\frac{g(\bm{y}_l)}{g(\bm{y}_w)}}_{\text{Adaptive Scaling}} \cdot \underbrace{(1 - \sigma(z))}_{\text{Confidence Weight}} \cdot \left( \underbrace{\frac{1}{|\bm{y}_l|} \nabla_\theta \log \pi_\theta(\bm{y}_l|\bm{x}) - \frac{1}{|\bm{y}_w|} \nabla_\theta \log \pi_\theta(\bm{y}_w|\bm{x})}_{\text{Direction of Policy Update}} \right).
\]

\textbf{Adaptive Scaling} \(\frac{g(\bm{y}_l)}{g(\bm{y}_w)}\):  
  For minimization problems, \(g(\bm{y}_w) < g(\bm{y}_l)\), so \(\frac{g(\bm{y}_l)}{g(\bm{y}_w)} > 1\). This amplifies the gradient magnitude for pairs where \(\bm{y}_l\) is significantly worse than \(\bm{y}_w\), prioritizing updates that correct large suboptimalities.

\textbf{Confidence Weight} \(1 - \sigma(z)\):  
  As the policy becomes more confident in preferring \(\bm{y}_w\) over \(\bm{y}_l\) (\(\sigma(z) \to 1\)), the gradient diminishes. This prevents overfitting to already well-separated pairs.

\textbf{Normalization} by \(|\bm{y}|\):  
  The normalization \(\frac{1}{|\bm{y}|}\) ensures that solutions of different lengths contribute equally to the gradient. Without this, longer solutions (e.g., JSP schedules with more operations) would dominate updates.

\textbf{Direction of Update}:  
  The gradient increases the likelihood of \(\bm{y}_w\) (since \(\nabla_\theta \log \pi_\theta(\bm{y}_w|\bm{x})\) is added) and decreases the likelihood of \(\bm{y}_l\) (since \(\nabla_\theta \log \pi_\theta(\bm{y}_l|\bm{x})\) is subtracted).

Compared with DPO and SimPO, their gradients are as follows:
\[
\nabla_\theta \mathcal{L}_{\text{DPO}} = -\beta \cdot \underbrace{(1 - \sigma(d))}_{\text{Confidence Weight}} \cdot \left( \underbrace{ \nabla_\theta \log \pi_\theta(\bm{y}_l|\bm{x}) -  \nabla_\theta \log \pi_\theta(\bm{y}_w|\bm{x})}_{\text{Direction of Policy Update}} \right),
\]

\[
\nabla_\theta \mathcal{L}_{\text{SimPO}} = -\beta \cdot \underbrace{(1 - \sigma(s))}_{\text{Confidence Weight}} \cdot \left( \underbrace{ \frac{1}{|\bm{y}_l|} \nabla_\theta \log \pi_\theta(\bm{y}_l|\bm{x}) - \frac{1}{|\bm{y}_w|} \nabla_\theta \log \pi_\theta(\bm{y}_w|\bm{x})}_{\text{Direction of Policy Update}} \right),
\]

where \[d=\beta\left(\log\frac{\pi_{\bm\theta}(\bm{y}_w|\bm x)}{\pi_{\text{ref}}(\bm{y}_w|\bm x)} - \log\frac{\pi_{\bm\theta}(\bm{y}_l|\bm x)}{\pi_{\text{ref}}(\bm{y}_l|\bm x)} \right), \;\; s=\beta\left(\frac{1}{|\bm{y}_w|}\log\frac{\pi_{\bm\theta}(\bm{y}_w|\bm x)}{\pi_{\text{ref}}(\bm{y}_w|\bm x)} - \frac{1}{|\bm{y}_l|}\log\frac{\pi_{\bm\theta}(\bm{y}_l|\bm x)}{\pi_{\text{ref}}(\bm{y}_l|\bm x)}-\frac{\gamma}{\beta} \right),  \]
represent the confidence weight in DPO and SimPO. 

DPO and SimPO rely on fixed hyperparameters (i.e., $\beta$ and $\gamma$) to control the gradient magnitude, requiring manual tuning and lacking dynamic adaptability. Furthermore, DPO does not perform length normalization on the log-likelihood calculations, which may lead to instability in scenarios with variable-length outputs. In contrast, {  BOPO} enhances robustness across diverse scenarios through adaptive scaling factors and length normalization, thereby reducing dependence on the $\beta$ hyperparameter.

\section{Runtime Analysis for JSP}
\label{ap:time}
We additionally provide the solving time in \cref{tb:time}, which is an important aspect in some scheduling scenarios. 

Our model, MGL, achieves significantly lower solving time compared to all non-constructive methods while delivering performance closing to L2S$_{5k}$. When compared to RL-based greedy constructive methods, MGL maintains competitive solving time even when sampling 512 solutions. Notably, MGL's solving time does not increase significantly with larger $B'$, a clear distinction from GAT-MHA, the model proposed in SLIM. This is because GAT-MHA relies on MHA decoder, which has a computational complexity of $O(n^2)$, a drawback that becomes particularly pronounced as $B'$ grows.

\begin{table}[b!]
\centering
    \caption{The average solving time (s) of algorithms on the TA benchmark. }
    \centering
    \small
    \label{tb:time}
\vskip 0.15in
\begin{tabular}{l||cccc|cccc|cc|cc}
\toprule \midrule
    & \multicolumn{4}{c|}{Non-constructive} & \multicolumn{4}{c|}{Greedy Constructive} & \multicolumn{2}{c}{$B'$=128}  & \multicolumn{2}{c}{$B'$=512}  \\ \cmidrule(lr){2-5}\cmidrule(lr){6-9}\cmidrule(lr){10-11}\cmidrule(lr){12-13}
Shape & OR-Tools & L2S$_{500}$ & TGA$_{500}$ & L2S$_{5k}$ & PDRs     & L2D      & SchN    & CL      & SLIM & {  BOPO} & SLIM & {  BOPO} \\ \midrule \toprule
15x15  & 462      & 9.3     & 12.6    & 92.2   & 0.00     & 0.39     & 3.5     & 0.80    & 0.69           & 0.67       & 0.72           & 0.75       \\
20x15  & 2880     & 10.1    & 14.6    & 102    & 0.00     & 0.60     & 6.6     & 1.10    & 0.84           & 0.80       & 1.07           & 0.96       \\
20x20  & 3600     & 10.9    & 17.5    & 114    & 0.00     & 0.72     & 11      & 1.39    & 1.11           & 1.06       & 1.37           & 1.23       \\
30x15  & 3600     & 12.7    & 17.2    & 120    & 0.01     & 0.95     & 17.1    & 1.49    & 1.24           & 1.19       & 1.83           & 1.44       \\
30x20  & 3600     & 14      & 19.3    & 144    & 0.01     & 1.41     & 28.3    & 1.72    & 1.66           & 1.59       & 2.42           & 1.91       \\
50x15  & 3600     & 16.2    & 23.9    & 168    & 0.01     & 1.81     & 52.5    & 2.82    & 2.19           & 1.99       & 4.06           & 2.60       \\
50x20  & 3600     & 22.8    & 24.4    & 228    & 0.02     & 3.00     & 96      & 3.93    & 2.91           & 2.63       & 5.41           & 3.44       \\
100x20 & 3600     & 50.2    & 42.0    & 504    & 0.19     & 9.39     & 444     & 9.58    & 7.85           & 5.31       & 20.05          & 7.96      \\
\midrule \bottomrule
\end{tabular}
\vskip -0.1in
\end{table}

\begin{figure}[b!]
\vskip 0.2in
    \centering
    \begin{subfigure}[b]{0.45\textwidth}
        \includegraphics[width=\textwidth]{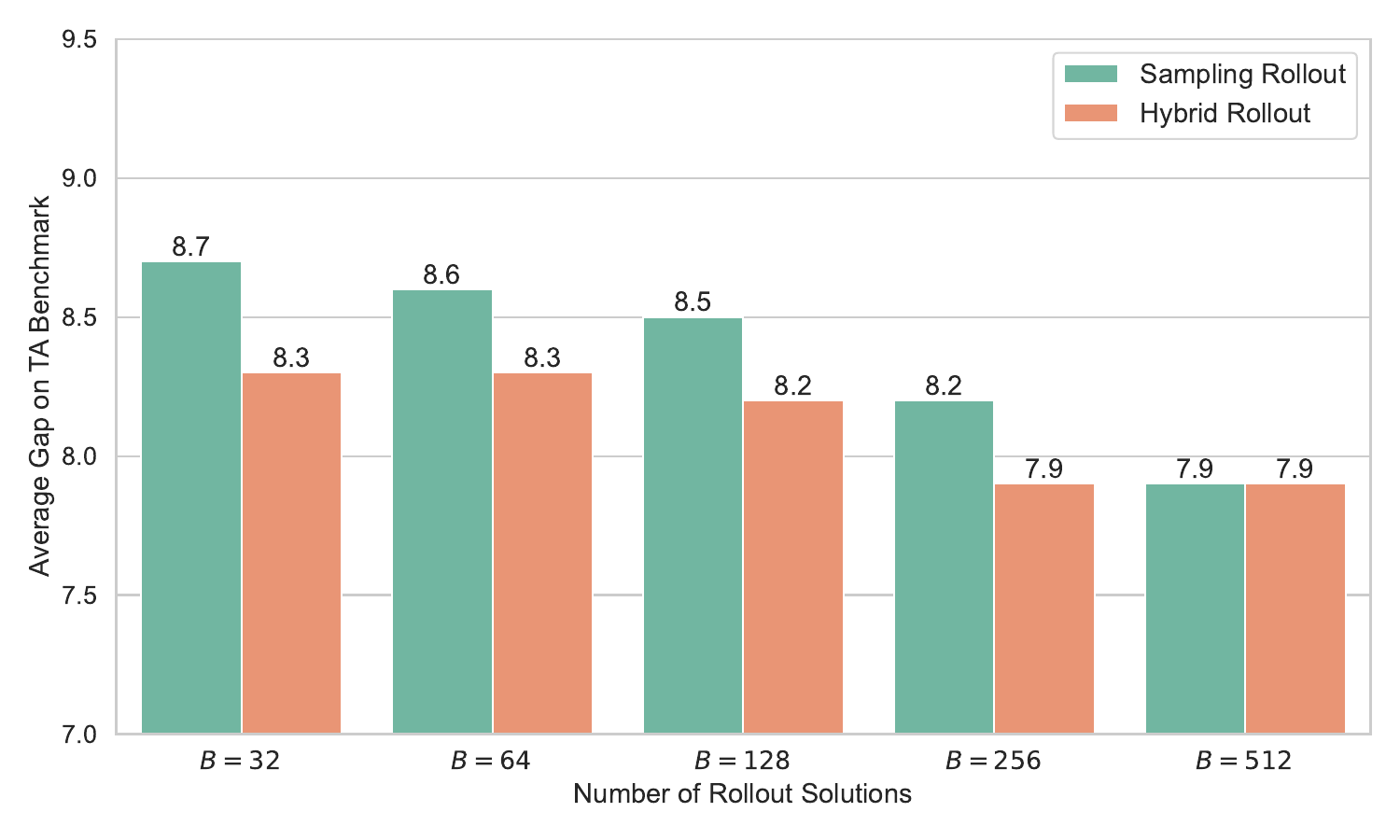} 
        \caption{}
        \label{fg:hybrid_rollout}
    \end{subfigure}
    \hfill
    \begin{subfigure}[b]{0.45\textwidth}
        \includegraphics[width=\textwidth]{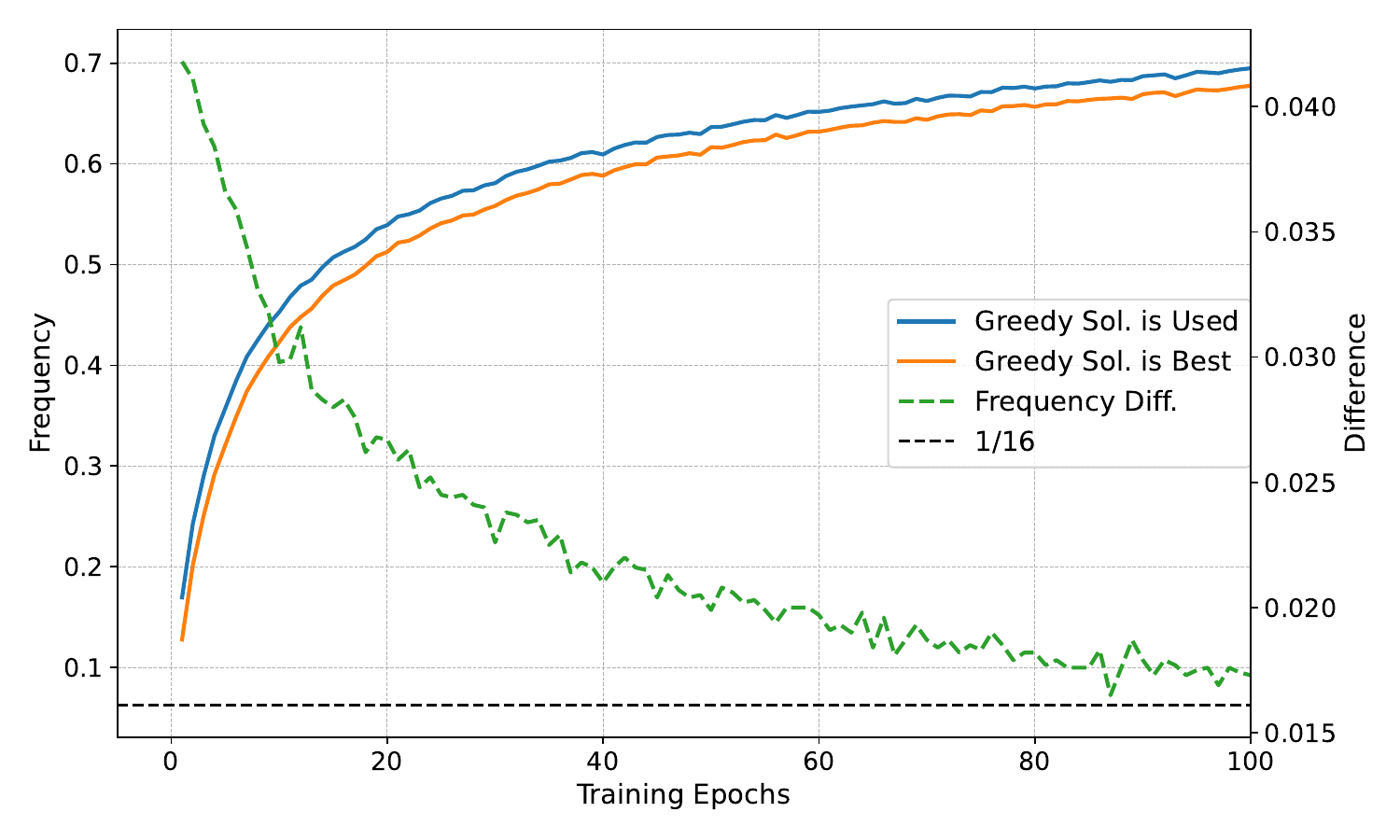}
        \caption{}
        \label{fg:greedy}
    \end{subfigure}
    \caption{Analysis of Hybrid Rollout. (a) Average gap (\%) on TA benchmark of different rollout methods with varying numbers of generated solutions $B$, (b) Participation and optimality of greedy solution during training.}
\vskip -0.2in
\end{figure}

\section{Further Analysis of Hybrid Rollout}
\label{ap:hybrid} 

To investigate the impact of introducing a greedy solution through hybrid rollout, we trained the model using both hybrid rollout and pure sampling approaches under different numbers of generated solutions $B$. The comparison results are shown in \cref{fg:hybrid_rollout}. Evidently, hybrid rollout not only improves model performance but also significantly reduces the dependency on $B$. Notably, the model achieves competitive results even with smaller generated sizes $B$. For instance, $B=32$ with hybrid rollout performs similarly to $B=64$ with hybrid rollout and is only 0.4 points behind $B=512$ with hybrid rollout. In contrast, $B=32$ without hybrid rollout performs poorly, with a gap of 0.8 compared to $B=512$ without hybrid rollout. 
Notably, hybrid rollout enables a generated solutions size of 256 to match the performance of a generated solutions size of 512 without hybrid rollout, demonstrating equivalent efficacy with reduced computational demand.

As illustrated in \cref{fg:greedy}, to investigate how greedy solutions enhance training efficacy, we trained the model on TSP20 with hyperparameters $B=128$ and $K=8$, systematically tracking two metrics per epoch:
\begin{itemize}
    \item The frequency of greedy solutions selected for preference pair construction (theoretical random selection probability: $1/16$);
    \item The frequency of greedy solutions being identified as the best solution.
\end{itemize}
The rising selection frequency of greedy solutions demonstrates their growing dominance in training. Given that greedy solutions exhibit the highest log-likelihood among all candidates, their participation delivers stronger gradient signals, accelerating model convergence.
The synchronized increase in greedy solutions being recognized as the best solution, and the decrease of difference between the frequencies mentioned above validates their critical role in improving solution set quality. This inherently addresses the limitation of sampling rollout, where suboptimal exploration often fails to capture high-quality candidates.

These demonstrate that introducing a greedy solution via hybrid rollout has an effect akin to increasing the solution generated size $B$ during training, but without incurring additional computational costs. This is particularly significant because large-scale and complex COPs often require substantial resources. Hybrid rollout offers a memory-efficient alternative while maintaining high performance.

\section{Additional Experiments.}
\label{ap:tsp}

{  To further investigate the impact of $B$ on different problem types, we evaluate our method on TSP20/50 with $B=20/50$ (denoted as BOPO$^-$). As shown in \cref{tb:more_tsp}, when sampling the same number of solutions as POMO, the performance of the model remains comparable to $B=128$. This suggests that for small-scale or simple problems, the optimal $B$ is low, as the model can efficiently sample high-quality solutions with fewer times.

BOPO's advantage primarily stems from pairwise preference learning, rather than simply being an RL variant with the proposed filtering method. To demonstrate this point, we compare BOPO with a POMO variant employing our Hybrid Rollout and Uniform Filtering (denoted as POMO$^+$). The key difference lies in the loss: while BOPO uses pairwise loss, POMO$^+$ calculates a standard POMO loss based on the filtered $K$ solutions, as it lacks the pairwise mechanism. Results in \cref{tb:more_tsp} show that POMO$^+$ is inferior to BOPO, demonstrating the effectiveness of the preference learning of BOPO. Moreover, POMO$^+$ is even inferior to POMO, underscoring that the filtering method is tightly adapted to BOPO and may not be applicable to general RL methods.
}

\begin{table}[htb!]
\centering
\small
\caption{Average gaps (\%) on generated TSP instances. }
\label{tb:more_tsp}
\vskip 0.15in
\begin{tabular}{l|l|l}
\toprule \midrule
Methods  & TSP20            & TSP50             \\ \midrule
POMO     & 0.002            & 0.042             \\
POMO$^+$  & 0.005            & 0.081             \\ \midrule
BOPO$^-$  & \textbf{0.000}   & \textbf{0.009}    \\
BOPO     & \textbf{0.000}   & \textbf{0.009}    \\
\bottomrule
\end{tabular}
\vskip -0.1in
\end{table}


\end{document}